\documentclass[fleqn, 10pt]{wlscirep}
\usepackage[utf8]{inputenc}
\usepackage[T1]{fontenc}
\usepackage{newunicodechar}
\usepackage{colortbl}
\usepackage{bm}
\usepackage{marvosym}
\usepackage{multirow} 
\usepackage{caption}
\usepackage{lineno}

\usepackage[ruled,vlined]{algorithm2e}
\usepackage{amsmath, amssymb}
\SetKwInput{KwInput}{Input}
\SetKwInput{KwOutput}{Output}

\title{Accurate and Scalable Multimodal Pathology Retrieval via Attentive Vision-Language Alignment}

\author[1,2]{Hongyi Wang*}
\author[1,3]{Zhengjie Zhu*}
\author[1]{Junlin Hou*}
\author[1]{Jiabo Ma}
\author[4]{Fang Wang}
\author[5]{Yue Shi}
\author[6]{Qiuyu Cai}
\author[6]{Jili Wang}
\author[7]{Bo Luo}
\author[1]{Zhizhong Chai}
\author[8]{Zhengyu Zhang}
\author[8]{Li Liang}
\author[6]{Xiuming Zhang}
\author[9 \Letter]{Yen-Wei Chen}
\author[2,10 \Letter]{Lanfen Lin}
\author[1,11,12,13,14 \Letter]{Hao Chen}

\affil[1]{Department of Computer Science and Engineering, The Hong Kong University of Science and Technology, Hong Kong SAR, China}
\affil[2]{College of Computer Science and Technology, Zhejiang University, Hangzhou, China.}
\affil[3]{School of Medicine, Wake Forest University, Winston-Salem, NC, USA.}
\affil[4]{Department of Radiology, Union Hospital, Tongji Medical College, Huazhong University of Science and Technology, Wuhan, China}
\affil[5]{Department of Pathology, Sir Run Run Shaw Hospital, School of Medicine, Zhejiang University, Hangzhou, China}
\affil[6]{Department of Pathology, The First Affiliated Hospital, School of Medicine, Zhejiang University, Hangzhou, China.}
\affil[7]{Department of Pathology, The Central Hospital of Wuhan, Tongji Medical College, Huazhong University of Science and Technology, Wuhan, China}
\affil[8]{Department of Pathology, Nanfang Hospital, School of Medicine, Southern Medical University, Guangzhou, China}
\affil[9]{College of Information Science and Engineering, Ritsumeikan University, Osaka, Japan.}
\affil[10]{Zhejiang Key Laboratory of Multi-omics Precision Diagnosis and Treatment of Liver Diseases, Zhejiang University, Hangzhou 310063, China}
\affil[11]{Department of Chemical and Biological Engineering, The Hong Kong University of Science and Technology, Hong Kong SAR, China}
\affil[12]{Division of Life Science, The Hong Kong University of Science and Technology, Hong Kong SAR, China}
\affil[13]{Shenzhen-Hong Kong Collaborative Innovation Research Institute, The Hong Kong University of Science and Technology, Shenzhen, China}
\affil[14]{State Key Laboratory of Nervous System Disorders, The Hong Kong University of Science and Technology, Hong Kong SAR, China}
\affil[*]{These authors contributed equally to this work.}
\affil[\Letter]{\textbf{Corresponding authors: Yen-Wei Chen (chen@is.ritsumei.ac.jp), Lanfen Lin (llf@zju.edu.cn), Hao Chen (jhc@cse.ust.hk)}}

\begin{abstract}

The rapid digitization of histopathology slides has opened new opportunities for computational tools in clinical and research workflows. Content-based slide retrieval can help pathologists identify morphologically and semantically related precedent cases, supporting expert diagnosis and example-based education. Effective retrieval of whole-slide images (WSIs), however, remains challenging because gigapixel slides contain abundant irrelevant content, focal diagnostic patterns and slide-level semantic information that must be represented at a practicable search cost. Here we present PathSearch, a retrieval framework that combines fine-grained attentive mosaics with slide-level embeddings aligned through vision-language contrastive learning. Trained on 6,926 slide-report pairs, PathSearch captures both fine-grained morphological cues and high-level semantic patterns to enable accurate and flexible retrieval. The framework supports two key functionalities: (1) mosaic-based image-to-image (I2I) retrieval, ensuring accurate and efficient slide search; and (2) multimodal retrieval, where text queries can directly retrieve relevant slides. PathSearch was evaluated on eight tasks comprising 5,021 evaluation slides, spanning malignancy assessment on frozen and hematoxylin and eosin (H\&E)-stained slides, lymph-node metastasis detection, tumor subtyping, mixed-gallery rare-cancer retrieval, and hepatocellular carcinoma (HCC) risk stratification. Internal and external experimental results demonstrate that PathSearch consistently outperforms the strongest existing methods without compromising multimodal accuracy. A multi-center reader study further demonstrated increases in task-level mean diagnostic accuracy, confidence, and inter-observer agreement with PathSearch's support. Together, these results support the effectiveness of PathSearch across diverse retrieval tasks and evaluation settings.

\end{abstract}

\begin{document}

    \maketitle

\begin{figure*}
\centering
\includegraphics[width=\textwidth]{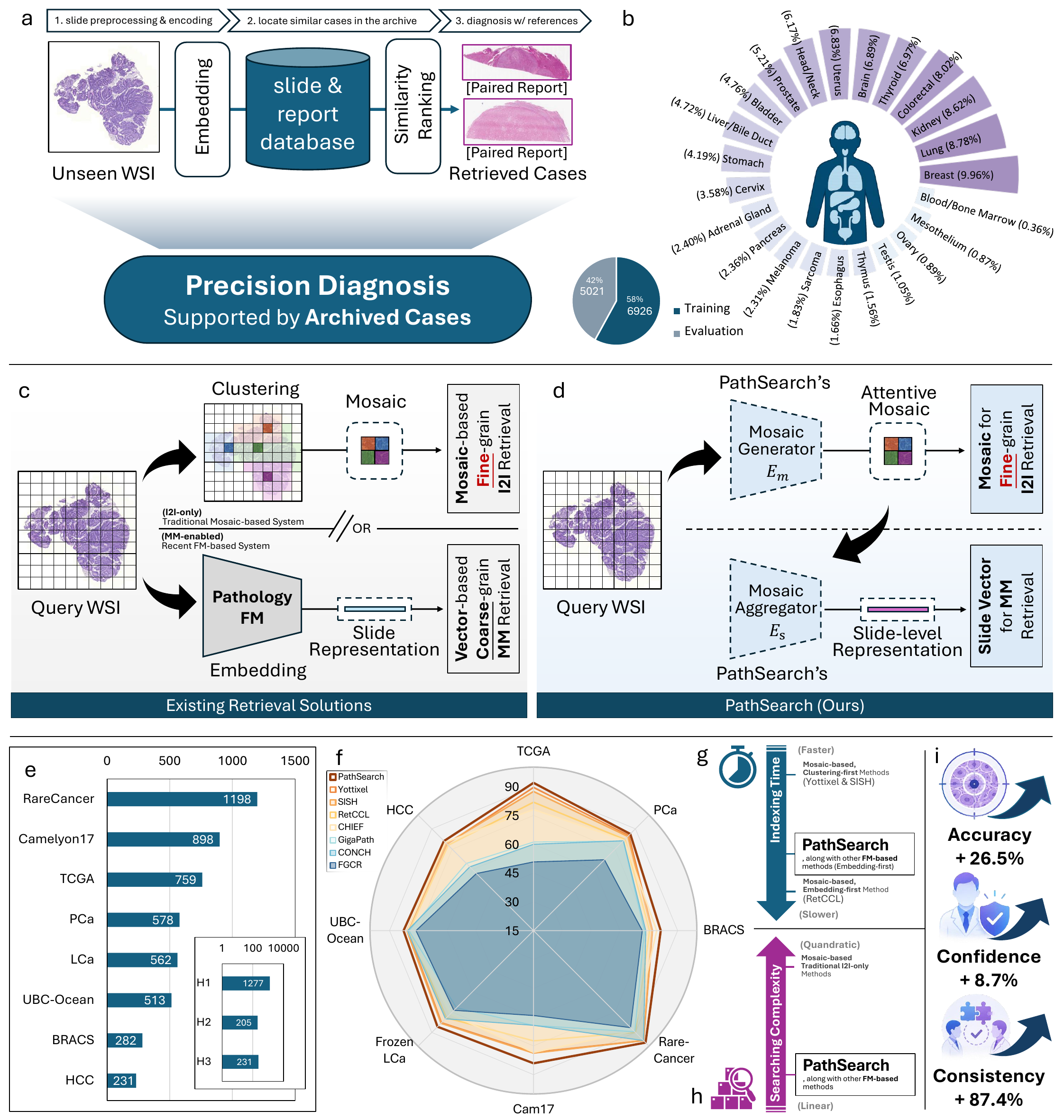}
\caption{\textbf{Overview of the PathSearch framework and retrieval pipeline.} 
\textbf{(a)} Clinical motivation: an unseen whole-slide image (WSI) is embedded and compared against a database of historical cases, enabling retrieval of precedent slides and reports to support precision diagnosis. 
\textbf{(b)} Composition of the study datasets, consisting of 6,926 slide-report training pairs from The Cancer Genome Atlas (TCGA) across diverse organs, and 5,021 evaluation slides across eight public and in-house tasks. 
\textbf{(c)} Existing retrieval systems are either mosaic-based image-to-image (I2I)-only systems, or foundation model (FM)-based multimodal systems that lack fine-grained visual similarity comparison.   
\textbf{(d)} PathSearch's workflow enables both mosaics for fine-grained I2I retrieval and slide-level embedding for multimodal retrieval.  
\textbf{(e)} WSI counts for different downstream tasks and in-house cohorts. 
\textbf{(f)} Radar plot summarizing comparative I2I retrieval performance across the eight evaluation tasks (Top-5 Majority Vote).
\textbf{(g)} The indexing time of PathSearch is in line with existing multimodal retrieval systems. 
\textbf{(h)} PathSearch has a linear searching complexity, which is more scalable than traditional I2I retrieval systems. 
\textbf{(i)} Across four readers, macro-averaged diagnostic accuracy increased from 68.0\% to 86.0\%, confidence from 7.77 to 8.44, and Fleiss' $\kappa$ from 0.33 to 0.63 with PathSearch.}
\label{overview}
\end{figure*}

\section{Introduction}\label{sec1}

The growing adoption of digital pathology has created large archives of whole-slide images (WSIs) that contain substantial diagnostic and morphological information \cite{bera2019artificial, pantanowitz2018twenty, moxley2020artificial, su2025computational}. Content-based image retrieval (CBIR) offers a complementary paradigm to predictive artificial intelligence by identifying previously diagnosed cases with similar morphology or pathology-report descriptions \cite{shmatko2022artificial, chen2022deep, yottixel}. Rather than providing only a predicted diagnostic label, retrieval systems enable pathologists to directly examine relevant reference cases and integrate this evidence with their own expertise (Figure~\ref{overview}(a)). Such case-based comparison may be particularly useful for diagnostically challenging cases and can also support pathology education and retrospective research.

Recent studies have demonstrated the feasibility of CBIR for diagnostic pathology across multiple disease settings \cite{yottixel, sish, retccl, fgcr, tizhoosh2024validation}. By linking retrieval results to interpretable reference cases, these systems provide an intuitive means of inspecting the evidence underlying image similarity. Human-centered evaluations have further suggested that pathologists value retrieval interfaces that allow interactive exploration and user guidance \cite{cai2019human}. However, translating this paradigm to large WSI archives remains challenging because WSIs are gigapixel-scale images with pronounced intra-slide morphological heterogeneity \cite{shao2021transmil, lin2024prompt, gpfm}. Existing approaches have largely followed two paradigms as shown Figure~\ref{overview}(c): mosaic-based image-to-image (I2I) retrieval and foundation model (FM)-based multimodal retrieval. Although each addresses an important aspect of WSI search, neither fully captures the requirements of accurate and flexible pathology retrieval.

Mosaic-based approaches reduce the computational burden of WSI retrieval by decomposing each slide into patches and selecting or clustering a subset of representative patches to approximate its morphological content \cite{yottixel, sish, retccl}. Frameworks such as Yottixel \cite{yottixel}, SISH \cite{sish}, and RetCCL \cite{retccl} have shown promising performance for visual similarity search \cite{lahr2024analysis}. However, their reliance on predefined or heuristic mosaic construction can constrain how slide-level morphology is represented and may limit correspondence with clinically meaningful concepts \cite{jahanifar2025domain, guo2025bpmambamil}. Moreover, because these methods primarily operate in the visual domain, they do not naturally support retrieval using free-text pathology descriptions or diagnostic queries \cite{cai2019human}.

On the other hand, recent pathology FMs have enabled multimodal retrieval by aligning visual and textual representations, allowing WSIs to be queried using natural-language descriptions, diagnostic terms or clinical concepts \cite{huang2024dynamic, huang2023language, he2024foundation, prism, prism2, mstar, musk}. Most of these approaches, however, follow CLIP-like architectures \cite{clip, plip} in which a WSI is represented by a single global embedding. Although such representations are well suited to semantic cross-modal matching, compressing an entire slide into one vector can obscure fine-grained morphological variation and local tissue patterns that are important for precise I2I matching. This creates an inherent trade-off between multimodal semantic alignment and preservation of detailed morphological information.

To address this trade-off, we developed PathSearch, a scalable multimodal retrieval framework that combines fine-grained morphological matching with slide-level vision-language alignment (Figure~\ref{overview}(d)). PathSearch replaces heuristic mosaic construction with a fixed set of attention-derived mosaic tokens that capture diverse and multi-scale morphological patterns for I2I retrieval. A mosaic aggregator subsequently integrates these tokens into a slide-level representation that is contrastively aligned with curated pathology reports, enabling image-to-image, image-to-text (I2T), text-to-image (T2I) and text-to-text (T2T) retrieval within a shared representation space. Because each WSI is represented by a fixed number of tokens irrespective of slide size, I2I retrieval retains linear exhaustive-search complexity with respect to database size, facilitating application to large pathology archives.

We evaluated PathSearch using a patient-level-audited benchmark from The Cancer Genome Atlas (TCGA) and multiple independent external cohorts encompassing cancer detection, tumor subtyping, rare-cancer retrieval and risk-label retrieval. Across these tasks, PathSearch achieved the highest top-5 majority-vote performance while maintaining practical indexing time and linear retrieval complexity (Figure~\ref{overview}(f-h)). In a reader study, access to PathSearch increased diagnostic accuracy, diagnostic confidence and inter-pathologist agreement by 26.5\%, 8.7\% and 87.4\%, respectively (Figure~\ref{overview}(i)). Together, these results support PathSearch as a general multimodal retrieval framework that combines detailed morphological comparison with natural-language search and may provide interpretable case-based support for diagnostic pathology.

\section{Results}\label{sec2}

\subsection{Data Characteristics}

An overview of the PathSearch retrieval system and study design is presented in Figure~\ref{overview}. As shown in Figure~\ref{overview}(b), a total of 6,926 unique slides from the publicly available TCGA \cite{tcga, tcga2} were used for training, with each slide paired with its corresponding pathology report. The training dataset spans multiple organ systems, including breast, lung, kidney, brain, prostate, colon, and gastric tissues, providing a morphologically and diagnostically diverse corpus for model development and subsequent evaluation.

Figure~\ref{overview}(e) summarizes the datasets used for downstream evaluation. PathSearch was evaluated across eight tasks comprising a total of 5,021 slides. The internal TCGA evaluation cohort contained 759 slides, including 257 slides for internal validation and 502 slides for held-out internal testing. The 502-slide test set covered six TCGA projects: lung adenocarcinoma (LUAD), lung squamous cell carcinoma (LUSC), kidney chromophobe (KICH), kidney renal papillary cell carcinoma (KIRP), kidney renal clear cell carcinoma (KIRC), and breast invasive carcinoma (BRCA). BRCA was further divided into invasive ductal carcinoma (IDC) and invasive lobular carcinoma (ILC) for retrieval evaluation. The TCGA dataset was partitioned at the patient level; because each patient contributed a single slide, this design also ensured the absence of slide-level overlap between the training, validation, and test partitions.

The seven external evaluation tasks comprised H1 prostate cancer malignancy classification (PCa; 578 slides), H1 frozen lung cancer malignancy classification (LCa; 562 slides), Camelyon17 lymph node metastasis classification \cite{camelyon17} (898 slides), UBC-OCEAN subtyping \cite{ubcocean} (513 slides), BRACS atypical-versus-malignancy classification \cite{brancati2022bracs} (282 slides), H1+H2 rare-cancer retrieval using a mixed gallery (1,198 slides), and H3 hepatocellular carcinoma (HCC) risk evaluation (231 slides) based on Edmondson-Steiner (ES) grading (i.e., ES I\&II are summarized as low risk, while ES III\&IV are high risk). The contributing in-house cohorts comprised 1,277 slides from H1, 205 slides from H2, and 231 slides from H3. Together with the 759-slide TCGA evaluation cohort, these task-level datasets comprised 5,021 evaluation slides. The H1+H2 rare-cancer retrieval task was evaluated in a query-gallery setting with distractor cases injected into the gallery, whereas all other tasks followed a leave-one-out retrieval paradigm.

\begin{figure*}
\centering
\includegraphics[width=\textwidth]{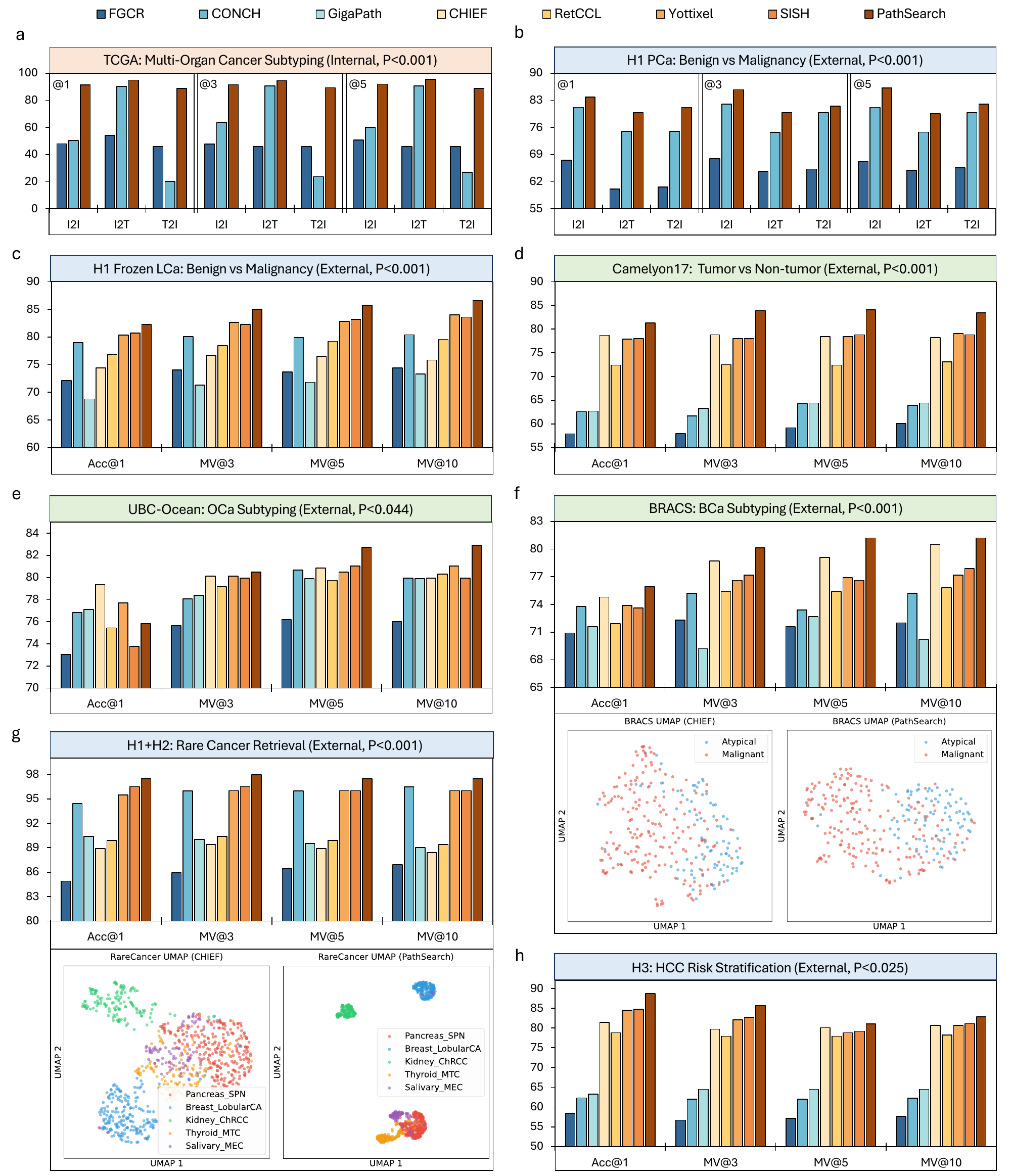}
\caption{\textbf{Performance of PathSearch and comparison methods across internal and external retrieval tasks.}
\textbf{(a)} Multimodal retrieval performance on the held-out internal TCGA test set comprising 502 slides from 502 patients, evaluated for I2I, I2T, and T2I retrieval.
\textbf{(b)} Multimodal retrieval performance on the external H1 PCa task for I2I, I2T, and T2I retrieval; corresponding T2T results are reported in Extended Data Table~\ref{tab:h1_pca_multimodal}.
\textbf{(c-h)} I2I retrieval performance on \textbf{(c)} H1 frozen LCa, \textbf{(d)} Camelyon17, \textbf{(e)} UBC-OCEAN, \textbf{(f)} BRACS (atypical versus malignant), \textbf{(g)} rare-cancer retrieval using a mixed gallery, and \textbf{(h)} H3 HCC risk evaluation. Reported $P$ values indicate paired, two-tailed t-tests based on nDCG@10 values of PathSearch and the best baseline within each task.}
\label{overall_results}
\end{figure*}

\subsection{PathSearch Outperforms State-of-the-Art Methods}

\noindent \textbf{Compared Methods.}
To evaluate retrieval performance, PathSearch was benchmarked against established WSI CBIR methods, including Yottixel \cite{yottixel}, SISH \cite{sish}, RetCCL \cite{retccl}, and FGCR \cite{fgcr}. We additionally compared PathSearch with the open-source pathology FMs CONCH \cite{conch}, GigaPath \cite{gigapath}, and CHIEF \cite{chief}, which use cosine similarity for retrieval. TITAN \cite{titan} was not included in the benchmarks because of its substantial computational requirements for slide-level indexing. In our experiments, TITAN required up to 8$\times$ longer indexing time than PathSearch. Its slide encoder relies on dense token-level attention, resulting in quadratic memory and computational complexity with respect to the number of patches. For 40$\times$ cohorts, such as H1, H3, and UBC-OCEAN, encoding a single slide with TITAN sometimes exceeded the 80\,GB GPU memory capacity of an NVIDIA A800 under FP16 inference. CPU-based indexing was substantially more resource-intensive, requiring 9.6-17+\,min and approximately 593\,GB of host memory per slide. We therefore excluded TITAN from the retrieval comparisons.

Among the compared methods, CONCH v1.5 was used as the common patch-level encoder for PathSearch, Yottixel, and SISH. This design minimizes differences arising from the underlying patch representations and allows for a more direct assessment of PathSearch's attentive mosaic mechanism. For RetCCL and FGCR, which incorporate their own specifically designed self-supervised patch encoders, we used the CCL \cite{retccl} and KAT \cite{kat} models, respectively, following their official implementations. For the FM-based methods, CONCH was evaluated by applying mean pooling over patch embeddings to obtain a slide-level representation because it was developed primarily as a patch-level pathology encoder. In contrast, CHIEF and GigaPath provide slide-level representations directly and were therefore applied to slide-to-slide retrieval without additional aggregation. As CHIEF and GigaPath were not trained with vision–language alignment, they were evaluated only for I2I retrieval and not for cross-modal retrieval. Together, these comparisons cover established mosaic-based retrieval systems, methods with specialized self-supervised encoders, and recent pathology FMs, while controlling the patch-level encoder where feasible.

\noindent \textbf{Evaluation Metrics.}
Following previous studies \cite{yottixel, sish, retccl, fgcr}, we evaluated retrieval performance using Top-1 Accuracy (Acc@1), Top-3 Majority Vote (MV@3), Top-5 Majority Vote (MV@5), and Top-10 Majority Vote (MV@10). For MV@$k$, the predicted category was defined as the most frequent label among the top-$k$ retrieved slides and was considered correct when it matched the ground-truth label of the query. When the top-$k$ retrieved results did not yield a unique majority label, the tie was resolved using the category of the highest-ranked retrieved slide. Normalized discounted cumulative gain (nDCG) \cite{ndcg} was additionally used in the reader study to quantify the ranking quality of the retrieved results. Computational efficiency was assessed through both theoretical complexity analysis with respect to slide size and database scale and empirical measurements of indexing time. Together, these metrics characterize retrieval accuracy, ranking quality, and computational scalability.

\begin{figure*}
    \centering
    \includegraphics[width=\textwidth]{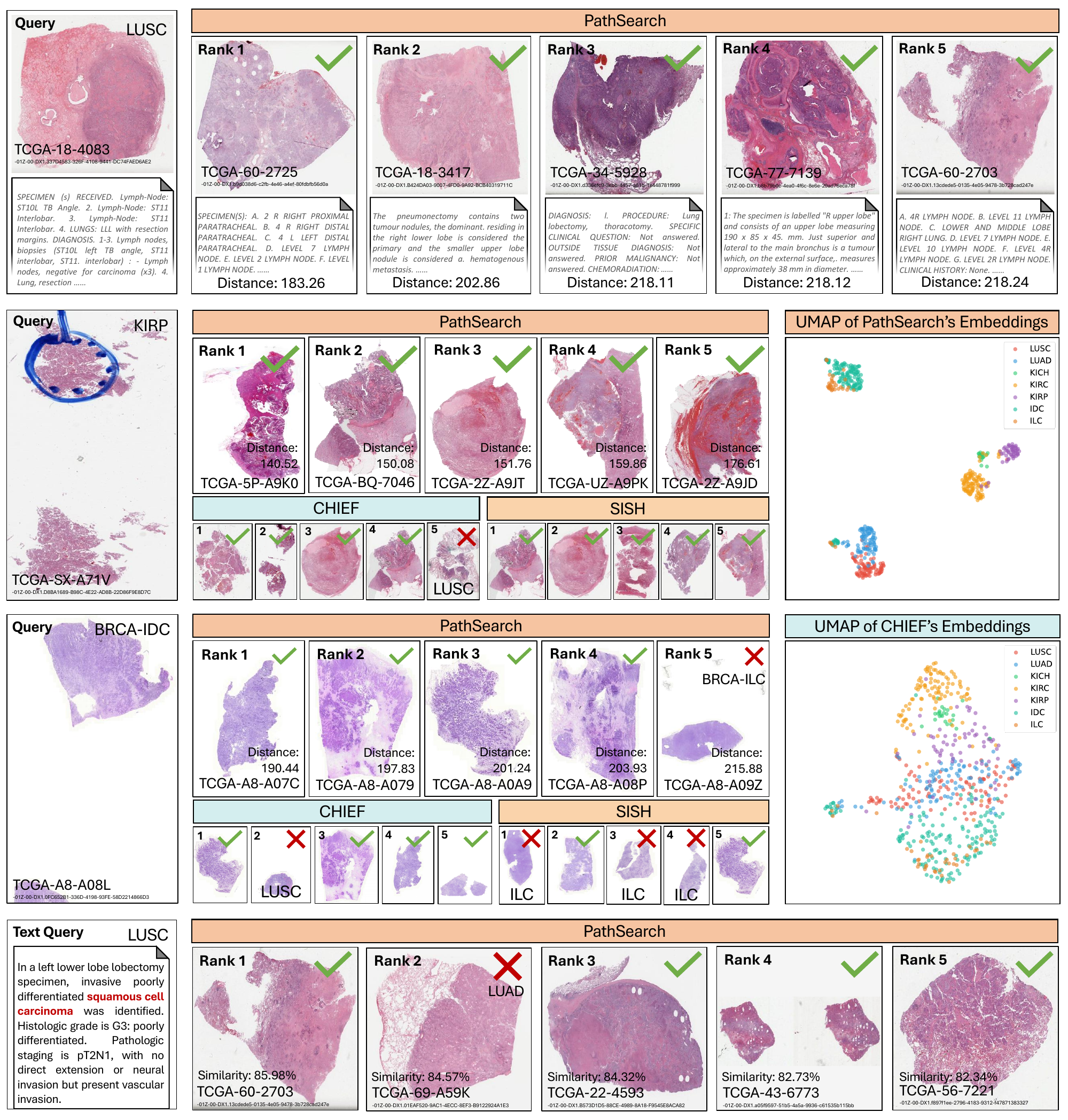}
    \caption{\textbf{Representative retrieval examples on the TCGA test set comparing PathSearch, CHIEF, and SISH.} Three image queries and one text query are shown together with their top-5 retrieved results. Green checkmarks (\checkmark) indicate label-concordant retrievals, whereas red crosses (\textcolor{red}{$\times$}) indicate label-discordant retrievals. \textbf{(Top row)} For a LUSC image query, PathSearch retrieves five LUSC slides. \textbf{(Second row)} For a KIRP query containing manual markings, PathSearch and SISH retrieve five KIRP slides, whereas CHIEF retrieves one LUSC slide among its top-5 results. \textbf{(Third row)} For a BRCA-IDC query, PathSearch retrieves four IDC slides and one ILC slide; the comparison methods show the corresponding retrieved results. The plots on the right show the embedding distributions generated by PathSearch and CHIEF for the selected examples. \textbf{(Bottom row)} For a text query describing LUSC, PathSearch retrieves four LUSC slides and one LUAD slide. These examples illustrate retrieval behavior across organ-level, subtype-level, and text-based queries and complement the quantitative evaluation.}
    \label{TCGA_Visualization}
\end{figure*}

\noindent \textbf{Overall Quantitative Results.}
The overall retrieval performance across internal and external evaluation tasks is summarized in Figure~\ref{overall_results}. These benchmarks span multiple organs, institutions, acquisition settings, and levels of diagnostic granularity, thereby testing retrieval performance under both cohort and task variation. Across the I2I dataset-metric combinations, PathSearch achieved the highest performance consistently, with the only exception being Acc@1 on UBC-OCEAN, for which CHIEF achieved 79.4\%, compared with 77.7\% for Yottixel and 75.8\% for PathSearch. On the same dataset, PathSearch achieved the highest performance for MV@3, MV@5, and MV@10, reaching 80.5\%, 82.7\%, and 82.9\%, respectively. Across all I2I comparisons, PathSearch improved over the strongest baseline method by a mean absolute margin of 2.1 percentage points. For the comparisons annotated in Figure~\ref{overall_results}, paired two-tailed t-tests of query-level nDCG@10 between PathSearch and the best-performing baseline yielded $P<0.001$ for six tasks, $P<0.044$ for UBC-OCEAN, and $P<0.025$ for H3 HCC. Collectively, these results indicate consistent retrieval performance across a diverse set of evaluation settings.

On the internal TCGA test set, which comprises multiple organ and cancer subtypes including closely related lung and breast categories, PathSearch achieved an Acc@1 of 91.4\%, MV@3 of 91.6\%, and MV@5 of 92.0\% for I2I retrieval. These values exceeded the strongest non-PathSearch comparator by 4.1, 4.1, and 2.4 percentage points, respectively (Figure~\ref{overall_results}(a) and Extended Data Table~\ref{tab:tcga_img2img}). For cross-modal retrieval, PathSearch achieved 95.0\%, 94.6\%, and 95.6\% for I2T and 88.8\%, 89.4\%, and 88.7\% for T2I across Acc@1, MV@3, and MV@5, respectively. The corresponding T2T results were 92.4\%, 95.8\%, and 95.2\%, respectively, and were higher than those of the comparison methods (Extended Data Table~\ref{tab:tcga_crossmodal}). These results show that PathSearch maintained strong performance for both visual and cross-modal retrieval within the same representation framework; the contributions of individual model components are examined separately in the ablation analyses.

The external H1 PCa cohort provided an additional evaluation setting in which the same benign-versus-malignant retrieval task was queried using both images and text. For I2I retrieval, PathSearch achieved 83.9\%, 85.8\%, and 86.2\% for Acc@1, MV@3, and MV@5, respectively. Corresponding performance was 79.8\%, 79.8\%, and 79.6\% for I2T and 81.2\%, 81.5\%, and 82.0\% for T2I (Figure~\ref{overall_results}(b)). For T2T retrieval, PathSearch achieved 99.8\% across all three metrics (Extended Data Table~\ref{tab:h1_pca_multimodal}). Together, these results demonstrate consistent label-concordant retrieval across different query modalities in an independent external cohort.

\noindent \textbf{Qualitative Results.}
Figure~\ref{TCGA_Visualization} presents representative retrieval examples from the TCGA test set for PathSearch, CHIEF, and SISH. For the LUSC query in the top row, PathSearch retrieves five label-concordant LUSC slides. The second row shows a KIRP query containing prominent manual markings; PathSearch and SISH each retrieve five KIRP slides, whereas one of the five slides retrieved by CHIEF is LUSC. This example suggests that the retrieval results can remain concordant despite visually salient non-tissue markings. For the BRCA-IDC query in the third row, PathSearch retrieves four IDC slides and one ILC slide, illustrating residual confusion between morphologically related breast cancer subtypes. Similarly, for the text query describing LUSC in the bottom row, four of the five retrieved slides are LUSC and one is LUAD, indicating remaining ambiguity between related lung cancer categories. These examples are intended to illustrate representative retrieval behavior and should be interpreted together with the complete quantitative evaluation in Figure~\ref{overall_results}.

\subsection{PathSearch Generalizes Across Different External Evaluations}

To assess generalizability beyond the internal TCGA benchmark, we evaluated PathSearch across seven external tasks spanning different organs, diagnostic settings, institutions, and label granularities: H1 PCa, H1 Frozen LCa, Camelyon17, UBC-OCEAN, BRACS, H1+H2 RareCancer retrieval, and H3 HCC risk evaluation. These tasks encompass tumor versus non-tumor or benign versus malignant discrimination, tumor subtyping and grouping, mixed-gallery rare-cancer retrieval, and risk evaluation. Together, they provide complementary tests of retrieval performance under changes in tissue morphology, acquisition conditions, institutional source, and clinical label definition.

\begin{figure*}
    \centering
    \includegraphics[width=\textwidth]{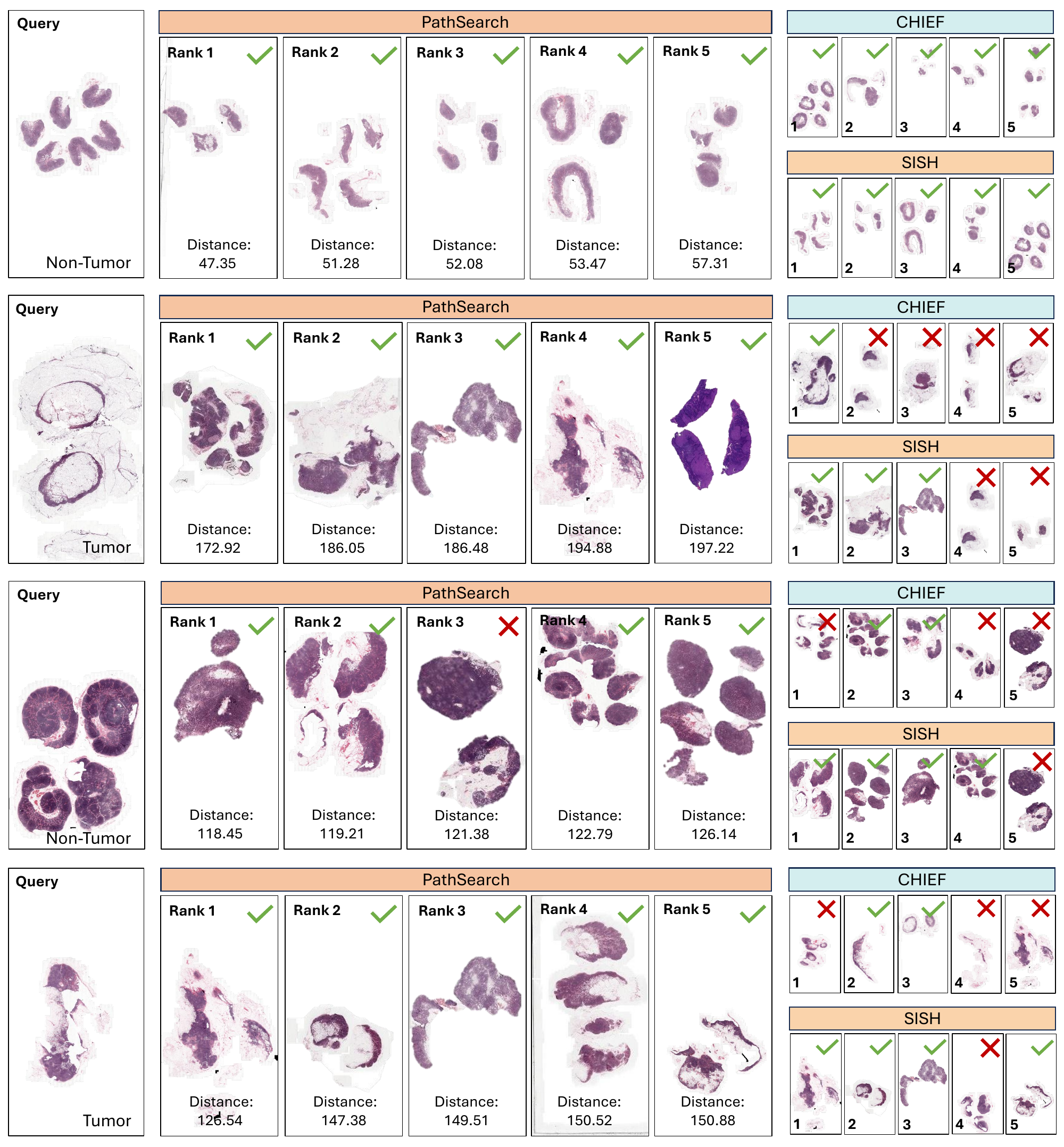}
    \caption{\textbf{Representative retrieval examples on the Camelyon17 dataset comparing PathSearch, CHIEF, and SISH.} Green checkmarks (\checkmark) indicate label-concordant retrievals, whereas red crosses (\textcolor{red}{$\times$}) indicate label-discordant retrievals. \textbf{(Top row)} For a non-tumor query, all three methods retrieve five non-tumor slides. \textbf{(Second row)} For a tumor query, PathSearch retrieves five tumor slides, whereas CHIEF and SISH retrieve multiple non-tumor slides. \textbf{(Third row)} For a more challenging non-tumor query, PathSearch and SISH each retrieve one label-discordant slide, whereas CHIEF retrieves three. \textbf{(Bottom row)} For a challenging tumor query, PathSearch retrieves five tumor slides, whereas CHIEF and SISH each include label-discordant results among their top-ranked retrievals. These examples illustrate retrieval behavior across queries with varying levels of difficulty and complement the aggregate quantitative evaluation.}
    \label{Cam17_Visualization}
\end{figure*}

\subsubsection{Tumor/Non-Tumor Diagnosis}

We first evaluated three external diagnostic-retrieval tasks: H1 PCa (578 slides), H1 Frozen LCa (562 slides), and Camelyon17 (898 slides). These tasks require retrieval of slides with concordant tumor status despite substantial background benign or non-tumor tissue, while H1 Frozen LCa and Camelyon17 additionally introduce variation related to tissue preparation and acquisition center.

On H1 PCa, PathSearch achieved 83.9\% Acc@1, 85.8\% MV@3, and 86.2\% MV@5 for I2I retrieval, exceeding the strongest non-PathSearch comparator by 1.2, 1.5, and 2.0 percentage points, respectively (Figure~\ref{overall_results}(b); Table~\ref{tab:h1_pca_img2img}). As described above, this cohort also supports multimodal retrieval, with corresponding I2T, T2I, and T2T results reported in Table~\ref{tab:h1_pca_multimodal}. These findings indicate that the performance advantage observed for I2I retrieval was retained across multiple retrieval cutoffs in this independent cohort.

Frozen-section preparation can introduce morphological artifacts that complicate matching between diagnostically similar slides. On H1 Frozen LCa, PathSearch achieved 82.3\% Acc@1, 85.0\% MV@3, 85.8\% MV@5, and 86.6\% MV@10 (Figure~\ref{overall_results}(c); Table~\ref{tab:frozen_lca}), exceeding the strongest non-PathSearch comparator by 1.5, 2.4, 2.5, and 2.6 percentage points, respectively. The consistent gains from Acc@1 through MV@10 indicate that the improvement extended beyond the highest-ranked retrieved slide to the broader retrieved set.

Camelyon17 evaluates lymph-node metastasis retrieval across multiple centers, introducing both institutional variation and substantial differences in tumor burden between slides. PathSearch achieved 81.3\% Acc@1, 83.9\% MV@3, 84.1\% MV@5, and 83.4\% MV@10 (Figure~\ref{overall_results}(d); Table~\ref{tab:camelyon17}), corresponding to improvements of 2.6, 5.1, 5.3, and 4.3 percentage points over the strongest non-PathSearch comparator, respectively. The larger gains for the majority-vote metrics than for Acc@1 indicate improved label consistency across multiple retrieved cases in this multi-center setting.

Representative Camelyon17 retrievals are shown in Figure~\ref{Cam17_Visualization}. The selected examples include queries for which all methods return label-concordant results, as well as more challenging cases in which tumor and non-tumor slides are mixed among the retrieved results. For the displayed challenging tumor query, PathSearch retrieves five label-concordant tumor slides, whereas the comparison methods include label-discordant results. These examples provide qualitative context for the aggregate results but are not intended to substitute for the complete quantitative evaluation.

\subsubsection{Tumor Subtyping and Grouping}

We next evaluated three external tasks involving tumor subtyping or diagnostic grouping: UBC-OCEAN (513 slides), BRACS (282 slides), and H1+H2 RareCancer retrieval (1,198 slides). These settings require retrieval across finer diagnostic distinctions than binary tumor-status classification and, in the RareCancer task, across a heterogeneous gallery containing multiple distractor organs and disease entities.

On UBC-OCEAN, which comprises morphologically related ovarian cancer subtypes, CHIEF achieved the highest Acc@1 at 79.4\%, whereas PathSearch achieved 75.8\% (Figure~\ref{overall_results}(e); Table~\ref{tab:ubc_ocean}). PathSearch nevertheless achieved the highest performance at the broader retrieval cutoffs, with MV@3, MV@5, and MV@10 values of 80.5\%, 82.7\%, and 82.9\%, respectively. These values exceeded the strongest non-PathSearch comparator by 0.4, 1.7, and 1.9 percentage points. Thus, although PathSearch did not achieve the highest nearest-neighbor accuracy on this task, it produced the most label-consistent retrieval sets at larger cutoffs, indicating more stable retrieval performance in this multi-class setting.

The BRACS atypical-versus-malignant task evaluates retrieval across a diagnostically subtle category boundary. PathSearch achieved 75.9\% Acc@1, 80.1\% MV@3, 81.2\% MV@5, and 81.2\% MV@10 (Figure~\ref{overall_results}(f); Table~\ref{tab:bracs}), exceeding the strongest non-PathSearch comparator by 1.1, 1.4, 2.1, and 0.7 percentage points, respectively. The improvements across all four retrieval cutoffs indicate consistently higher label concordance within this external breast pathology cohort.

The RareCancer task represents a distinct retrieval setting in which queries from five disease groups were searched against a mixed gallery comprising cases from H1 and H2 across multiple organs and morphologies. To increase task difficulty, the gallery was augmented with distractor slides, including normal tissue and tumors from non-target subtypes across the five organs. PathSearch achieved 97.5\% Acc@1, 98.0\% MV@3, 97.5\% MV@5, and 97.5\% MV@10 (Figure~\ref{overall_results}(g); Table~\ref{tab:rare_cancer}), corresponding to improvements of 1.0, 1.5, 1.5, and 1.0 percentage points over the strongest baseline comparator  of each metric, respectively. These results indicate that PathSearch maintained high retrieval accuracy despite the heterogeneous composition of the gallery.

\subsubsection{Tumor Grading and Risk Evaluation}

Finally, we evaluated PathSearch on the H3 HCC risk task comprising 231 slides. In contrast to organ-level or tumor-status retrieval, this task requires discrimination between risk labels within a single disease entity, where relevant morphological differences may be more subtle. PathSearch achieved 88.7\% Acc@1, 85.7\% MV@3, 81.0\% MV@5, and 82.8\% MV@10 (Figure~\ref{overall_results}(h); Table~\ref{tab:fahwmu_hcc_risk}), exceeding the strongest baseline of each metric by 3.9, 3.0, 0.9, and 1.7 percentage points, respectively. The improvements across all four retrieval cutoffs indicate that the performance advantage extended to retrieval based on clinically relevant risk stratification, although the magnitude of improvement varied across metrics.

\begin{figure*}
    \centering
    \includegraphics[width=\textwidth]{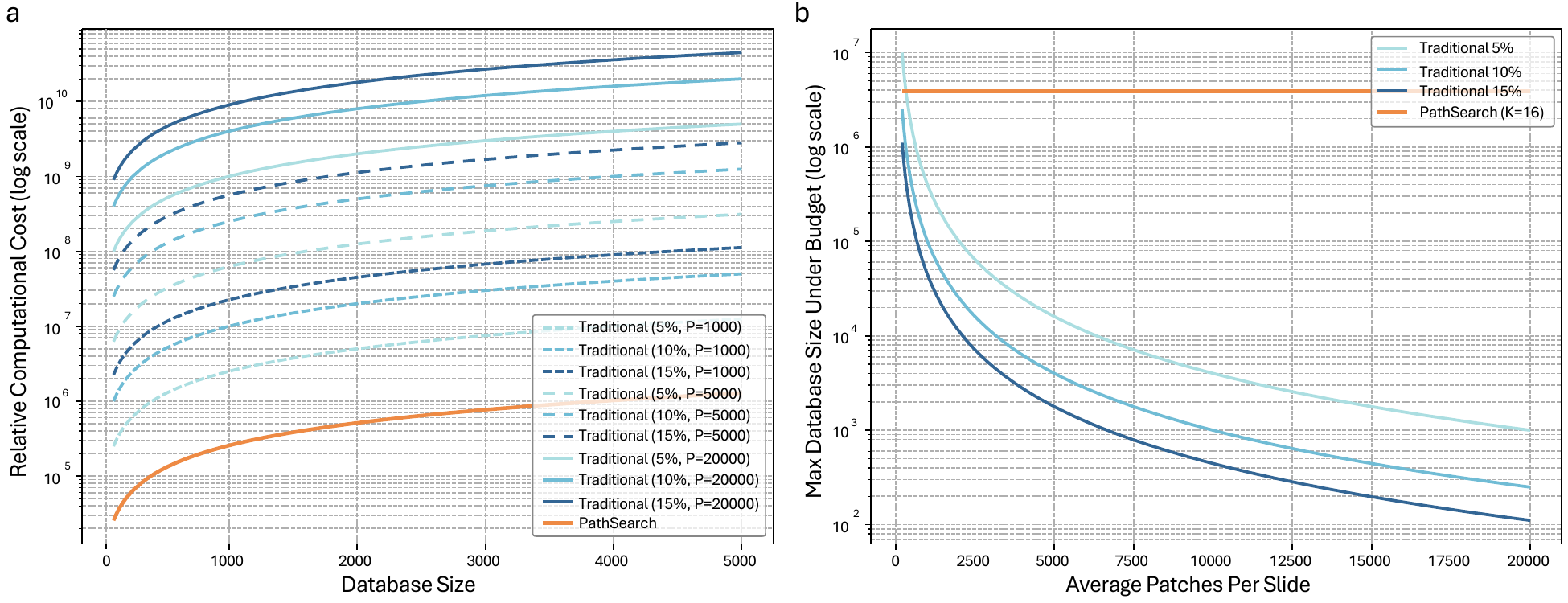}
    \caption{\textbf{Computational scalability of PathSearch compared with conventional mosaic-based retrieval methods.}
    Theoretical retrieval cost is analyzed as a function of database size and the number of tissue patches per slide.
    \textbf{(a)} Log-scale comparison of relative computational cost as the number of slides in the database, denoted by $S$, increases. PathSearch scales linearly with $S$ because each slide is represented by a fixed number of mosaic tokens. For conventional proportion-based mosaic methods, the per-candidate comparison cost additionally depends quadratically on the number of sampled patches. Curves are shown for sampling fractions $f$ of 5\%, 10\%, and 15\% and slide patch counts $P$ of $1,000$, $5,000$, and $20,000$.
    \textbf{(b)} Maximum database size supported under a fixed computational budget as a function of $P$. Because PathSearch uses a fixed-size mosaic, its theoretical capacity is independent of $P$ under the stated model, whereas the capacity of proportion-based mosaic methods decreases in proportion to $1/P^2$.}
    \label{complexity_comparison}
\end{figure*}

\subsection{PathSearch Remains Scalable for Large-Scale Databases}

Computational scalability is an important consideration for WSI retrieval, particularly as pathology archives increase in size. Retrieval efficiency is determined by both the cost of indexing new slides and the computational complexity of searching the resulting database. We therefore evaluated these two components separately through empirical indexing-time measurements and theoretical analysis of search complexity.

\subsubsection{Indexing Time}

Indexing time measures the computational cost of converting an input WSI into a database-ready representation. The evaluated indexing pipeline comprises patch tiling, patch-level feature extraction, and, where applicable, mosaic construction or feature aggregation. Because the tiling procedure was identical across all compared methods, its runtime was excluded from this analysis. To isolate differences attributable to mosaic construction and aggregation, CONCH was used as the patch-level encoder for all methods in this experiment. Detailed timing results are provided in Extended Data Table~\ref{indexing_time_comparison}.

The evaluated approaches can be broadly separated into clustering-first and embedding-first indexing strategies. Yottixel and SISH follow a clustering-first procedure in which patches are first grouped using low-level image features, such as color histograms, to identify representative regions. Patch-level embeddings are then computed only for the selected patches, while the remaining patches are excluded from the final mosaic. In our experiments, clustering required an average of 9.75\,s per slide when 10\% of patches were sampled for mosaic construction. Despite this additional clustering step, embedding only the selected patches reduced the overall indexing time to 47.51\,s per slide, approximately $2.5\times$ lower than that of the evaluated embedding-first approaches.

By contrast, RetCCL, FGCR, and FM-based approaches follow an embedding-first strategy in which features are extracted from all tissue patches before slide-level representation or mosaic construction. This increases the patch-embedding workload but allows subsequent aggregation to operate directly in the learned feature space. PathSearch similarly embeds all tissue patches and then constructs its representation through attentive aggregation. The aggregation step itself required only 0.0131\,s per slide, substantially less than the clustering time required by the clustering-first approaches. Consequently, the indexing cost of PathSearch is dominated by patch-level feature extraction rather than mosaic construction. Although its total indexing time is higher than that of clustering-first methods that embed only a subset of patches, it remains within the range of the evaluated embedding-first approaches while supporting the retrieval performance reported above.

\subsubsection{Search Scalability}

Search complexity determines how retrieval costs change as the database and individual slides increase in size. We compared PathSearch with two broad classes of retrieval approaches: conventional mosaic-based systems, including Yottixel, SISH, and RetCCL, and methods that perform retrieval using a single slide-level embedding, including FGCR, CONCH, and CHIEF. Detailed derivations are provided in Sec.~\ref{extended_complexity_reasoning}, and Figure~\ref{complexity_comparison} shows model-based comparisons under the corresponding complexity assumptions. Yottixel is plotted as a representative proportion-based mosaic method because methods using the same pairwise mosaic-comparison strategy exhibit the same asymptotic scaling behavior. Slide-embedding methods are not plotted separately because their exhaustive-search complexity scales linearly with database size, similar to PathSearch.

Conventional proportion-based mosaic methods retain a fraction $f$ of the tissue patches from each WSI. When retrieval involves pairwise comparisons between the sampled patches of the query and candidate slides, the per-candidate computational cost scales as $\mathcal{O}((fP)^2)$, where $P$ denotes the number of tissue patches per slide. The total exhaustive-search cost therefore scales as $\mathcal{O}(S(fP)^2)$ for a database containing $S$ slides. As shown in Figure~\ref{complexity_comparison}(a), this dependence on $P$ becomes increasingly consequential as slide resolution and tissue area increase. Under a fixed computational budget, the corresponding supportable database size consequently decreases with increasing $P$ (Figure~\ref{complexity_comparison}(b)).

PathSearch instead represents each slide using a fixed number of mosaic tokens. With $K=16$ tokens per slide, the cost of comparing a query with an individual candidate is $\mathcal{O}(K^2)$ and is therefore constant with respect to the original number of WSI patches. Because $K$ is fixed, exhaustive retrieval scales as $\mathcal{O}(S)$ with database size. This scaling behavior is comparable to methods based on a single slide-level embedding, while PathSearch additionally retains a token-level representation for fine-grained I2I comparison.

Together, these analyses characterize the computational trade-off of PathSearch. Its indexing procedure requires embedding all tissue patches and is therefore slower than clustering-first approaches that embed only a selected subset. However, the subsequent attentive aggregation adds negligible indexing overhead, and the fixed-size representation enables linear exhaustive-search complexity with database size without dependence on the original WSI patch count.

\begin{figure*}
\centering
\includegraphics[width=\textwidth]{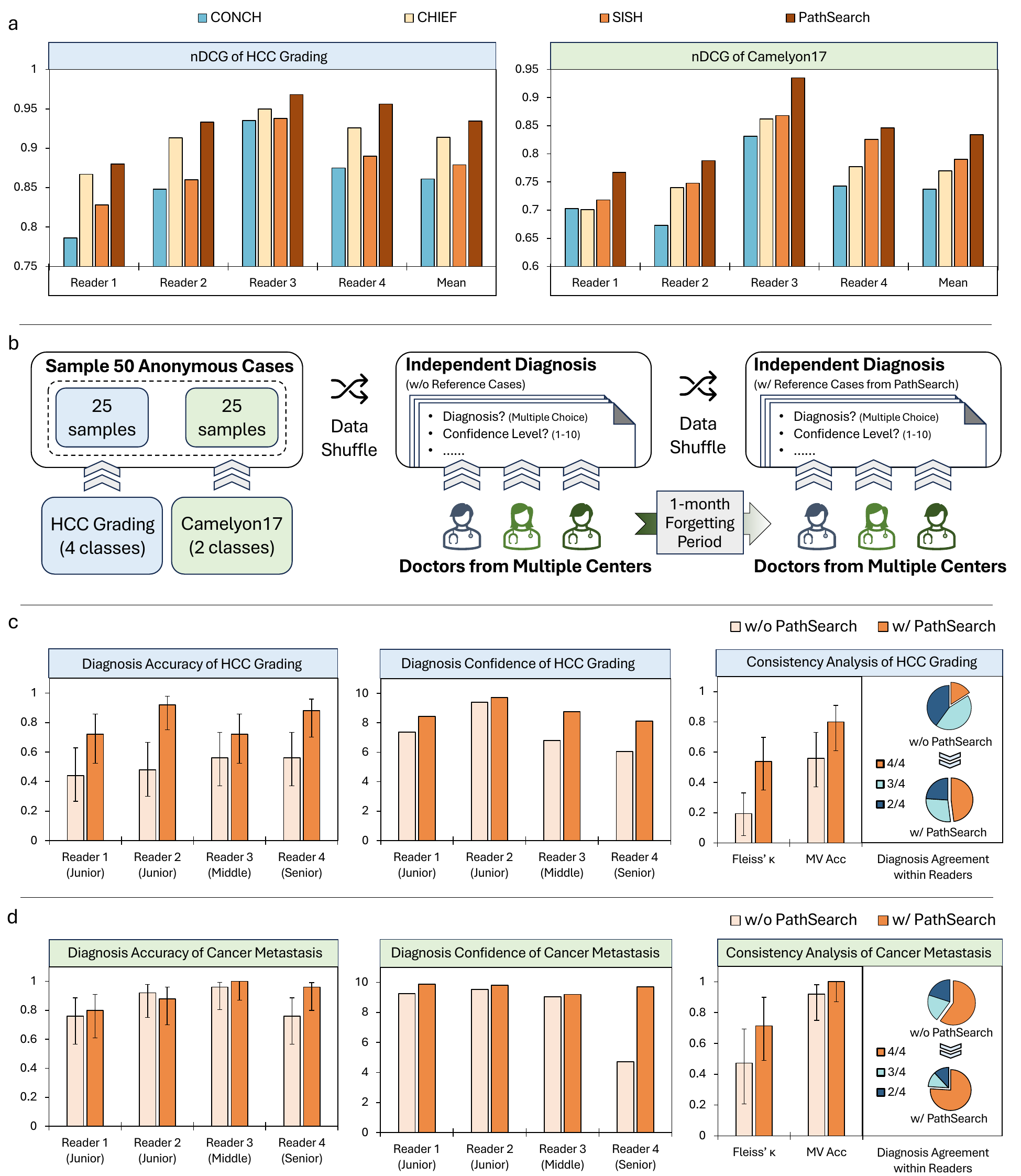}
\caption{\textbf{Reader-study evaluation of PathSearch across two diagnostic tasks.}
\textbf{(a)} nDCG@5 for four readers on HCC grading and Camelyon17, comparing the ranking quality of retrieval results generated by PathSearch and the comparison methods.
\textbf{(b)} Reader-study design. Pathologists from multiple centers first evaluated 50 anonymized cases independently and, after a 1-month washout period, reassessed the cases in a shuffled order with PathSearch retrieval support.
\textbf{(c)} Per-reader diagnostic accuracy and confidence with and without PathSearch support for HCC grading, together with Fleiss' $\kappa$, strict three-of-four reader consensus accuracy, and the distribution of reader agreement.
\textbf{(d)} Corresponding results for the Camelyon17 breast-cancer metastasis task.}
\label{HiSRA_reader_study}
\end{figure*}

\subsection{Reader Study Shows Task- and Reader-Dependent Effects}

To assess the potential clinical utility of PathSearch, we conducted a reader study involving four pathologists from three medical centers with different levels of experience: two early-career pathologists with 3 and 5 years of experience, one mid-career pathologist with 6 years of experience, and one senior pathologist with 16 years of experience. Each reader independently evaluated 50 de-identified slides, comprising 25 Camelyon17 slides for breast cancer lymph-node metastasis identification and 25 slides from the in-house H3 HCC cohort, which were evaluated as a more challenging four-class ES grading task for the pathologists. For each query, the top five retrieved slides were presented as reference cases for subjective nDCG@5 assessment and as diagnostic support during the reader study.

We first assessed the perceived relevance and ranking quality of the retrieved reference cases. Figure~\ref{HiSRA_reader_study}(a) shows reader-specific nDCG@5 values for PathSearch and the comparison methods across the two tasks, with the corresponding numerical results reported in Extended Data Table~\ref{tab:reader_ndcg}. PathSearch showed the strongest overall ranking performance across readers in both HCC grading and Camelyon17, providing evidence that its retrieved cases were ranked consistently with pathologist-assessed relevance.

The reader study subsequently evaluated whether access to PathSearch retrieval results was associated with changes in diagnostic performance (Figure~\ref{HiSRA_reader_study}(b)). In the first session, pathologists independently provided their diagnoses and confidence ratings without retrieval assistance. After a 1-month washout period to reduce recall effects, the same cases were reshuffled and reassessed with PathSearch support. This paired design enabled within-reader comparison of diagnostic accuracy, confidence, and agreement before and after retrieval support.

Figures~\ref{HiSRA_reader_study}(c) and~\ref{HiSRA_reader_study}(d) summarize the diagnostic results for the two tasks. For HCC grading, accuracy increased from 0.44 to 0.72 for the first early-career reader, from 0.48 to 0.92 for the second early-career reader, from 0.56 to 0.72 for the mid-career reader, and from 0.56 to 0.88 for the senior reader. Mean accuracy across the four readers consequently increased from 0.51 without retrieval support to 0.81 with PathSearch support. For breast-cancer metastasis detection, accuracy changed from 0.76 to 0.80, from 0.92 to 0.88, from 0.96 to 1.00, and from 0.76 to 0.96 for the four readers, respectively, resulting in an increase in mean accuracy from 0.85 to 0.91. Thus, the magnitude of the observed effect differed substantially across tasks and readers, with the largest gains occurring in the more challenging HCC grading task and one reader showing a modest decrease in accuracy on Camelyon17.

Self-reported diagnostic confidence increased for every reader-task combination. Mean confidence increased from 7.40 to 8.01 for HCC grading and from 8.13 to 8.87 for metastasis detection. The senior reader had the lowest baseline confidence on both tasks, increasing from 6.04 to 7.12 for HCC grading and from 4.72 to 6.60 for metastasis detection. These results indicate that retrieval support was consistently associated with higher self-reported confidence, although the magnitude of change varied among readers and tasks.

We further assessed inter-reader consistency using Fleiss' $\kappa$, strict three-of-four vote accuracy, and the distribution of reader agreement. Strict vote accuracy defined a case as correctly classified when at least three of the four pathologists provided the correct diagnosis, whereas the agreement distribution quantified the proportions of cases with 2/4, 3/4, or 4/4 reader agreement. For HCC grading, Fleiss' $\kappa$ increased from 0.196 to 0.538, majority vote accuracy from 0.56 to 0.80, and complete 4/4 agreement from 20\% to 48\%. For breast-cancer metastasis detection, Fleiss' $\kappa$ increased from 0.472 to 0.714, majority vote accuracy from 0.92 to 1.00, and 4/4 agreement from 60\% to 76\%. Together, these concordant changes across multiple agreement measures indicate greater inter-reader consistency when PathSearch retrieval results were available.

\section{Discussion}

In this study, we developed PathSearch, a multimodal CBIR framework that combines fine-grained attentive mosaic representations with slide-level vision-language alignment. Existing WSI retrieval systems typically prioritize either local morphological comparison or compact global representations, creating a trade-off between fine-grained visual matching, multimodal semantic retrieval, and computational scalability. PathSearch was designed to bridge these representations within a unified retrieval framework. Across the eight evaluation tasks, PathSearch achieved the highest overall performance while not compromising image- and text-based retrieval. The reader study further provided preliminary evidence that the retrieved reference cases were relevant to pathologists and that access to these cases was associated with improvements in mean diagnostic accuracy, confidence, and inter-reader agreement.

Several findings provide insight into the contribution of the proposed representation. First, across the eight ablation tasks, introducing the attentive mosaic mechanism yielded mean relative improvements of 1.6\%, 1.4\%, and 1.2\% for Acc@1, MV@3, and MV@5, respectively. Second, combining mosaic-level representations with slide-level semantic embeddings maintained strong performance across both nearest-neighbor and majority-vote retrieval in most external evaluations, suggesting that local morphological matching and global semantic information provide complementary retrieval signals. Third, alignment of slide representations with pathology reports enabled a shared representation space supporting image- and text-based queries on both TCGA and the external H1 PCa cohort. Together, these findings suggest that fine-grained visual representations and slide-level semantic alignment can be integrated without requiring separate retrieval systems for visual and textual queries.

The reader study also highlights both the potential and the context dependence of retrieval-assisted diagnosis. The largest increase in mean diagnostic accuracy was observed for HCC grading, where baseline reader performance and agreement were comparatively low, whereas improvements were smaller for Camelyon17, for which baseline performance was already high for some readers. One reader showed a modest reduction in metastasis-detection accuracy despite increased confidence, underscoring that retrieval support does not uniformly improve individual decisions. These findings suggest that the clinical value of case-based retrieval may depend on task difficulty, reader experience, and the quality and relevance of the retrieved references. Importantly, PathSearch is intended to provide reference evidence for pathologist interpretation rather than to replace the reader's diagnostic judgment.

Several limitations should be considered. First, PathSearch was trained on 6,926 slide-report pairs, which is substantially smaller than the training corpora used for some recent pathology FMs. Scaling training to larger and more heterogeneous collections may improve robustness, particularly for uncommon diseases and underrepresented tissue types. Second, the present study primarily considered hematoxylin and eosin (H\&E)-stained slides. Extending the framework to additional stains and complementary molecular modalities, including immunohistochemistry, spatial transcriptomics, or genomic data, could enable retrieval based on a broader range of clinically relevant information. Third, although evaluation included multiple external cohorts and institutions, the current reader study involved a limited number of four pathologists, 50 cases, and two diagnostic tasks. Larger prospective studies will be required to determine whether the observed changes in diagnostic performance translate to routine clinical workflows and to assess effects on diagnostic efficiency, error patterns, and user interaction.

PathSearch also introduces a modest storage overhead by retaining attentive mosaic tokens in addition to the slide-level semantic representation. In the current implementation, these tokens require approximately 1.5~KB per slide, equivalent to approximately one quarter of the storage required for a double-precision semantic vector. By comparison, a conventional mosaic retaining 5\% of the patches from a WSI containing 10,000 tissue patches can require up to approximately 48~KB per slide under the evaluated representation. Combined with the linear exhaustive-search complexity demonstrated above, this fixed-size representation provides a practical storage-computation trade-off for scaling retrieval to larger archives.

In summary, PathSearch provides a scalable multimodal approach to WSI retrieval that combines fine-grained morphological comparison with semantic supervision from pathology reports. It showed consistently strong retrieval performance across diverse organs, institutions, acquisition settings, and diagnostic tasks while supporting both image- and text-based queries. In the reader study, PathSearch support was associated with higher task-level mean diagnostic accuracy, greater self-reported confidence, and increased inter-reader agreement, with effects varying by task and reader. These findings support further investigation of case-based retrieval for pathology archive exploration, education, and diagnostic decision support, with prospective validation required before clinical deployment.


\section{Methods}

\subsection{Data Preparation}
\label{detailed_text_preprocessing}

Because the datasets included in this study were acquired using different scanning protocols and magnifications, preprocessing was adapted to the acquisition characteristics of each dataset. Within each dataset, however, identical preprocessing was applied to all compared methods. For TCGA, WSIs were tiled into non-overlapping $512\times512$ patches at level~0 from slides scanned at $40\times$ magnification for both model development and evaluation. External cohorts were processed according to their respective acquisition settings, with the same dataset-specific pipeline used for all methods evaluated on a given cohort.

For the in-house cohorts, slides from H1 and H2 were scanned using an SQARY-SQS-600P scanner, whereas slides from H3 were scanned using a KFBIO-400-PRO scanner. The corresponding cohorts contained 1,277, 205, and 231 slides from H1, H2, and H3, respectively.

Across all datasets, background and low-information regions were excluded using a tissue mask generated by Otsu's thresholding \cite{otsu}. Tiles containing less than 80\% foreground tissue were removed. For feature extraction, CONCH \cite{conch} was used to encode each retained patch into a 768-dimensional embedding, representing each WSI as a set of $N$ patch-level feature vectors.

To enable controlled comparison of slide-level retrieval strategies, CONCH was also used as the patch encoder for Yottixel and SISH. This configuration ensured that differences among PathSearch, Yottixel, and SISH primarily reflected their respective mosaic construction and retrieval mechanisms rather than differences in patch-level representations. RetCCL and FGCR, by contrast, incorporate patch encoders that constitute an integral component of their original methods; their corresponding pretrained encoders were therefore retained following the respective implementations. For slide-level FMs such as CHIEF, representations were generated directly using their original inference pipelines without additional training or modification.

For pathology-report curation, we developed a structured pipeline to convert raw TCGA diagnostic documents into text representations paired with their corresponding WSIs. Diagnostic reports were downloaded from the TCGA repository in PDF format and converted to text using optical character recognition (OCR), yielding 7,685 WSI-report pairs across multiple organs and disease subtypes. Of these, 6,926 pairs were assigned to the training cohort, whereas 257 and 502 slides from distinct patients were assigned to the internal validation and held-out test cohorts, respectively. The combined TCGA evaluation cohort therefore comprised 759 slides from 759 patients. Detailed cohort distributions by TCGA project are provided in Extended Data Table~\ref{tcga_training} and Table~\ref{tcga_testing}.

Because OCR conversion introduced recognition errors and formatting artifacts, Qwen3-Max was used to correct corrupted or inconsistent passages before subsequent text processing. The same model was then used to generate concise summaries for vision-language alignment. Summarization was designed to preserve slide-observable pathological findings while excluding information not directly inferable from histology, such as tumor location and tumor size. The following prompt was used:

\begin{quote}
``Summarize the following clinical report scanned from a PDF into NO MORE than 77 words. Avoid using bullet points. Remove tumor location and tumor size information. Keep the information about the pathological type of tumor, nuclear morphology, tumor grading, staining, tissue morphology, arrangement, and whether there is microvascular invasion if such information exists. Directly begin the summary with <summary> tag and end it with </summary> tag: [original report content]''
\end{quote}

To assess the quality of the resulting summaries, 500 cases were randomly sampled and independently reviewed again by pathologists. All the reviewed summaries were judged to correctly preserve the relevant pathological information from the source reports.

\subsection{Implementation Details}
\label{sec_implementation_details}

PathSearch was implemented in PyTorch 2.4.0 and trained on two NVIDIA A800 GPUs with 80\,GB memory each. Training was performed for 100 epochs with a batch size of 128 using the AdamW optimizer, an initial learning rate of $8\times10^{-5}$, and a weight decay of 0.05. Automatic mixed-precision training was enabled to reduce memory consumption and computational cost. The checkpoint with the best performance on the internal validation cohort was used for subsequent evaluation.

To prevent information leakage, the TCGA cohort was partitioned at the patient level according to a predefined split strategy. Because each patient contributed a single slide in the TCGA cohort used in this study, patient-level partitioning also ensured the absence of slide-level overlap among the training, validation, and test sets. For LUAD, LUSC, KIRP, KIRC, KICH, and BRCA, 257 slides from 257 patients were reserved for internal validation and 502 slides from 502 patients were reserved for held-out testing. The remaining samples from these projects, together with samples from the other TCGA cancer types, were used exclusively for model training, resulting in 6,926 training slides.

For contrastive learning, negative pairs were constructed within each mini-batch, such that each matched image-text pair was contrasted against all unmatched pairs in the same batch. Random seeds were fixed for all experiments to improve reproducibility.

\begin{figure*}
\centering
\includegraphics[width=\textwidth]{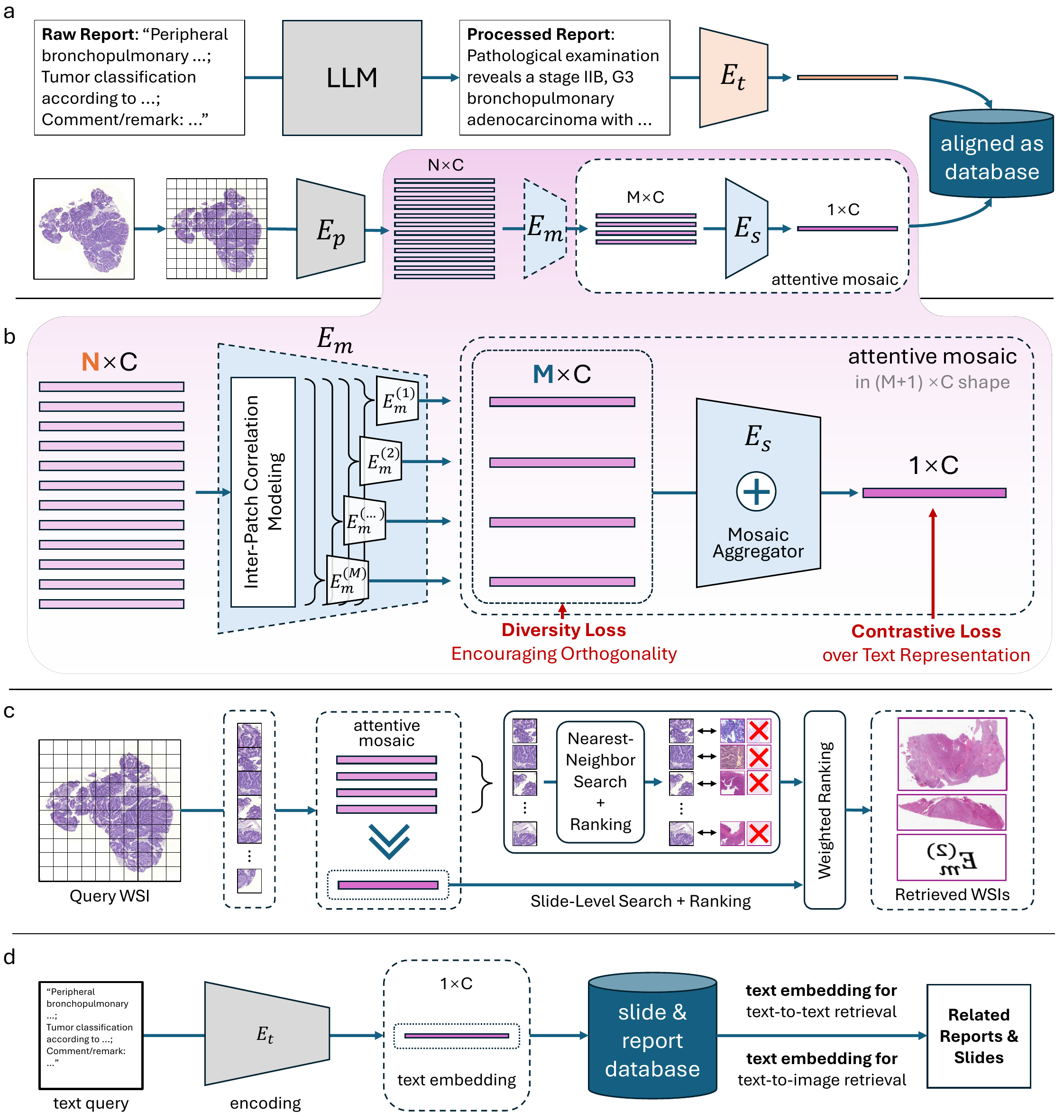}
\caption{\textbf{Architecture and retrieval procedure of PathSearch.}
\textbf{(a)} Training framework. Whole-slide images (WSIs) and reports curated using a LLM are encoded and aligned through contrastive learning.
\textbf{(b)} Vision branch. $N$ patch representations are processed through $M$ mosaic token generators to form the $M\times C$ attentive mosaic. These tokens are then aggregated into a slide-level $1\times C$ semantic representation through $E_s$. Training combines a diversity loss ($L_d$) and a contrastive loss ($L_c$) aligning slide representation with the report representation.
\textbf{(c)} I2I retrieval combines Hamming distances between binarized mosaic tokens with semantic distances between slide-level representations through weighted fusion.
\textbf{(d)} T2I and T2T retrieval use cosine similarity between text queries and slide or report representations in the shared embedding space.}
\label{detailed_pipeline}
\end{figure*}

\subsection{PathSearch Pipeline}

The overall training framework of PathSearch is illustrated in Figure~\ref{detailed_pipeline}(a). In the vision branch, each WSI is first encoded as a sequence of patch-level embeddings generated by a pretrained pathology FM $E_p$ (i.e., CONCH in our work). These patch embeddings are subsequently processed by a learnable attentive mosaic module $E_m$ that converts a variable-length set of patch representations into a fixed number of mosaic tokens. The mosaic representation accommodates variation in slide dimensions, tissue area, and the resulting number of patches while preserving multiple morphological patterns for I2I retrieval. After that, a mosaic aggregator $E_s$ further integrates these tokens into a single slide-level semantic embedding, which is aligned with the paired pathology-report representation through contrastive learning. During I2I retrieval, both the mosaic tokens and the aggregated slide-level embedding contribute to the retrieval distance. For cross-modal retrieval, the aligned slide-level embedding is used directly to retrieve the corresponding textual representations.

In the language branch, pathology reports are processed using the report-curation procedure described above, including OCR correction and summarization with the Large Language Model (LLM) Qwen-Max \cite{qwen3}. The 6,926 curated reports in the training cohort are subsequently encoded using a trainable text encoder $E_t$ initialized from PubMedBERT \cite{pubmedbert}. The report-level representation is obtained from the designated sequence-level output token of the text encoder and projected to the same dimensionality as the slide-level visual representation. The resulting visual and textual embeddings are optimized jointly in a shared representation space through contrastive learning, enabling retrieval across image and text modalities.

A more detailed illustration of the vision branch is shown in Figure~\ref{detailed_pipeline}(b). Given a patch-level embedding sequence of size $N\times C$, where $N$ denotes the number of tiled patches and $C$ the feature dimension, an inter-patch correlation module based on self-attention \cite{transformer} is first applied to incorporate contextual information across patch representations. The resulting features are then passed to an attentive mosaic generator $E_m$ that produces a fixed set of mosaic tokens.

Because $N$ varies substantially across WSIs, the attentive mosaic generator contains $M$ parallel attention branches, each generating one mosaic token. For the $m$-th branch, the corresponding token is computed as
\begin{equation}
    H_m = \sum_{n=1}^{N} a_{mn} h_n,
\end{equation}
where
\begin{equation}
    a_{mn}
    =
    \frac{
    \exp\left\{
    \omega_m^{T}
    \left[
    \tanh(V_{1,m}h_n)
    \odot
    \sigma(V_{2,m}h_n)
    \right]
    \right\}
    }
    {
    \sum_{j=1}^{N}
    \exp\left\{
    \omega_m^{T}
    \left[
    \tanh(V_{1,m}h_j)
    \odot
    \sigma(V_{2,m}h_j)
    \right]
    \right\}
    }.
\end{equation}
Here, $H_m$ denotes the $m$-th mosaic token, $h_n$ denotes the contextualized representation of the $n$-th patch, and $a_{mn}$ is the corresponding attention weight assigned by the $m$-th branch. The parameters $\omega_m$, $V_{1,m}$, and $V_{2,m}$ are independently learned for each attention branch, and $\odot$, $\tanh(\cdot)$, and $\sigma(\cdot)$ denote element-wise multiplication, the hyperbolic tangent function, and the sigmoid function, respectively. Because the attention operation accepts an arbitrary number of input instances, the resulting representation contains a fixed number $M$ of mosaic tokens irrespective of the original number of WSI patches. Unless otherwise specified, $M$ was set to 16 according to the parameter study described below.

The mosaic tokens are further aggregated into a slide-level semantic representation for alignment with the paired pathology report. The mosaic aggregator uses an attention-based operation to integrate information across the $M$ tokens. Model training combines a bidirectional contrastive loss $L_c$ for vision-language alignment with a diversity regularization term $L_d$ that reduces redundancy among the learned mosaic tokens. The overall objective is
\begin{equation}
    L = L_c + \alpha L_d,
\end{equation}
where $\alpha$ controls the contribution of the diversity term and was set to 1 unless otherwise specified.

For a mini-batch containing $B$ matched image-text pairs, the contrastive loss is defined as
\begin{equation}
    L_c =
    \frac{1}{2}
    \left(
    -\frac{1}{B}
    \sum_{i=1}^{B}
    \log
    \frac{\exp(s_{ii})}
    {\sum_{j=1}^{B}\exp(s_{ij})}
    -
    \frac{1}{B}
    \sum_{i=1}^{B}
    \log
    \frac{\exp(s_{ii})}
    {\sum_{j=1}^{B}\exp(s_{ji})}
    \right),
\end{equation}
with
\begin{equation}
    s_{ij} = \tau \cdot {E^v_i}^{T}E^t_j,
\end{equation}
where $E^v_i$ and $E^t_j$ denote the visual and textual embeddings, respectively, and $\tau$ denotes the learnable similarity-scaling parameter.

The diversity loss is computed across the $M$ mosaic tokens as
\begin{equation}
    L_d =
    \frac{1}{M^2-M}
    \sum_{\substack{i,j=1\\i\neq j}}^{M}
    c_{ij},
\end{equation}
where
\begin{equation}
    c_{ij} = {E^m_i}^{T}E^m_j,
\end{equation}
and $E^m_i$ denotes the embedding of the $i$-th mosaic token. This regularization discourages different attention branches from converging to highly similar representations and thereby reduces redundancy among the learned mosaic tokens.

The contrastive and diversity objectives are optimized jointly rather than sequentially: $L_d$ regularizes the set of mosaic tokens, whereas $L_c$ propagates semantic supervision from the paired reports through the aggregated slide representation to the vision branch. To stabilize optimization of the language encoder, only the final three Transformer blocks of PubMedBERT were fine-tuned, while the preceding layers remained frozen.

\begin{algorithm}[t]
\caption{PathSearch I2I Retrieval by Fusion of Mosaic and Semantic Distances}
\label{alg:hisra_i2i}
\KwInput{
Query slide $q$; candidate database $\mathcal{D}=\{s_i\}_{i=1}^{S}$; \\
Patch encoder $\phi_{\mathrm{patch}}$; attentive mosaic generator $\mathcal{M}$; \\
Mosaic aggregator $\psi$; retrieval-fusion weight $\beta \geq 0$.
}
\KwOutput{
Candidates ranked by fused distance $D_{\mathrm{fuse}}(q,s_i)$.
}

\BlankLine
\textbf{(1) Encode query slide}\;
Tile $q$ and obtain patch embeddings:
$\{\mathbf{z}^{q}_{n}\}_{n=1}^{N_q}
\leftarrow
\phi_{\mathrm{patch}}(q)$\;

Generate and binarize $M$ attentive mosaic tokens:
$\mathbf{B}_{q}
=
\{\mathbf{b}^{q}_{m}\}_{m=1}^{M}
\leftarrow
\mathrm{bin}
\left(
\mathcal{M}
\left(
\{\mathbf{z}^{q}_{n}\}_{n=1}^{N_q}
\right)
\right)$\;

Aggregate the mosaic tokens to obtain an L2-normalized slide-level semantic vector:
$\mathbf{v}_{q}
\leftarrow
\mathrm{norm}
\left(
\psi\left(
\mathcal{M}
\left(
\{\mathbf{z}^{q}_{n}\}_{n=1}^{N_q}
\right)
\right)
\right)$\;

\BlankLine
\textbf{(2) Load precomputed candidate representations}\;
\ForEach{$s_i \in \mathcal{D}$}{
    Retrieve the cached binarized mosaic
    $\mathbf{B}_{i}
    =
    \{\mathbf{b}^{i}_{m}\}_{m=1}^{M}$
    and normalized semantic representation $\mathbf{v}_{i}$\;
}

\BlankLine
\textbf{(3) Compute mosaic distance}\;
\ForEach{$s_i \in \mathcal{D}$}{
    $D_{\mathrm{mosaic}}(q,s_i)
    \leftarrow
    \mathrm{MedianMinHamming}
    (\mathbf{B}_{q},\mathbf{B}_{i})$\;
}

\BlankLine
\textbf{(4) Compute semantic distance}\;
\ForEach{$s_i \in \mathcal{D}$}{
    $D_{\mathrm{sem}}(q,s_i)
    \leftarrow
    \|\mathbf{v}_{q}-\mathbf{v}_{i}\|_2$\;
}

\BlankLine
\textbf{(5) Normalize distance scores}\;
Compute $\mu_{\mathrm{m}}$ and $\sigma_{\mathrm{m}}$ over
$\{D_{\mathrm{mosaic}}(q,s_i)\}_{i=1}^{S}$,
and $\mu_{\mathrm{s}}$ and $\sigma_{\mathrm{s}}$ over
$\{D_{\mathrm{sem}}(q,s_i)\}_{i=1}^{S}$\;

\ForEach{$s_i \in \mathcal{D}$}{
    $\tilde{D}_{\mathrm{mosaic}}(q,s_i)
    \leftarrow
    \dfrac{
    D_{\mathrm{mosaic}}(q,s_i)-\mu_{\mathrm{m}}
    }{
    \sigma_{\mathrm{m}}+\varepsilon
    }$\;

    $\tilde{D}_{\mathrm{sem}}(q,s_i)
    \leftarrow
    \dfrac{
    D_{\mathrm{sem}}(q,s_i)-\mu_{\mathrm{s}}
    }{
    \sigma_{\mathrm{s}}+\varepsilon
    }$\;
}

\BlankLine
\textbf{(6) Fuse distances and rank candidates}\;
\ForEach{$s_i \in \mathcal{D}$}{
    $D_{\mathrm{fuse}}(q,s_i)
    \leftarrow
    \tilde{D}_{\mathrm{mosaic}}(q,s_i)
    +
    \beta\,
    \tilde{D}_{\mathrm{sem}}(q,s_i)$\;
}

Return candidates in ascending order of $D_{\mathrm{fuse}}(q,s_i)$\;

\end{algorithm}

\begin{algorithm}[t]
\caption{MedianMinHamming Distance Between Attentive Mosaics}
\label{alg:median_min_hamming}
\KwInput{
Query mosaic $\mathbf{B}_q=\{\mathbf{b}^{q}_{m}\}_{m=1}^{M}$;
candidate mosaic $\mathbf{B}_i=\{\mathbf{b}^{i}_{m'}\}_{m'=1}^{M}$.
}
\KwOutput{
Mosaic distance $D_{\mathrm{mosaic}}(q,s_i)$.
}

\For{$m=1$ \KwTo $M$}{
    Compute
    $H_{m,m'}
    \leftarrow
    \mathrm{Hamming}
    (\mathbf{b}^{q}_{m},\mathbf{b}^{i}_{m'})$
    for all $m'\in\{1,\ldots,M\}$\;

    $d_m
    \leftarrow
    \min_{m'}
    H_{m,m'}$\;
}

Return
$D_{\mathrm{mosaic}}(q,s_i)
\leftarrow
\mathrm{median}
\left(
\{d_m\}_{m=1}^{M}
\right)$\;

\end{algorithm}

\begin{figure*}
\centering
\includegraphics[width=\textwidth]{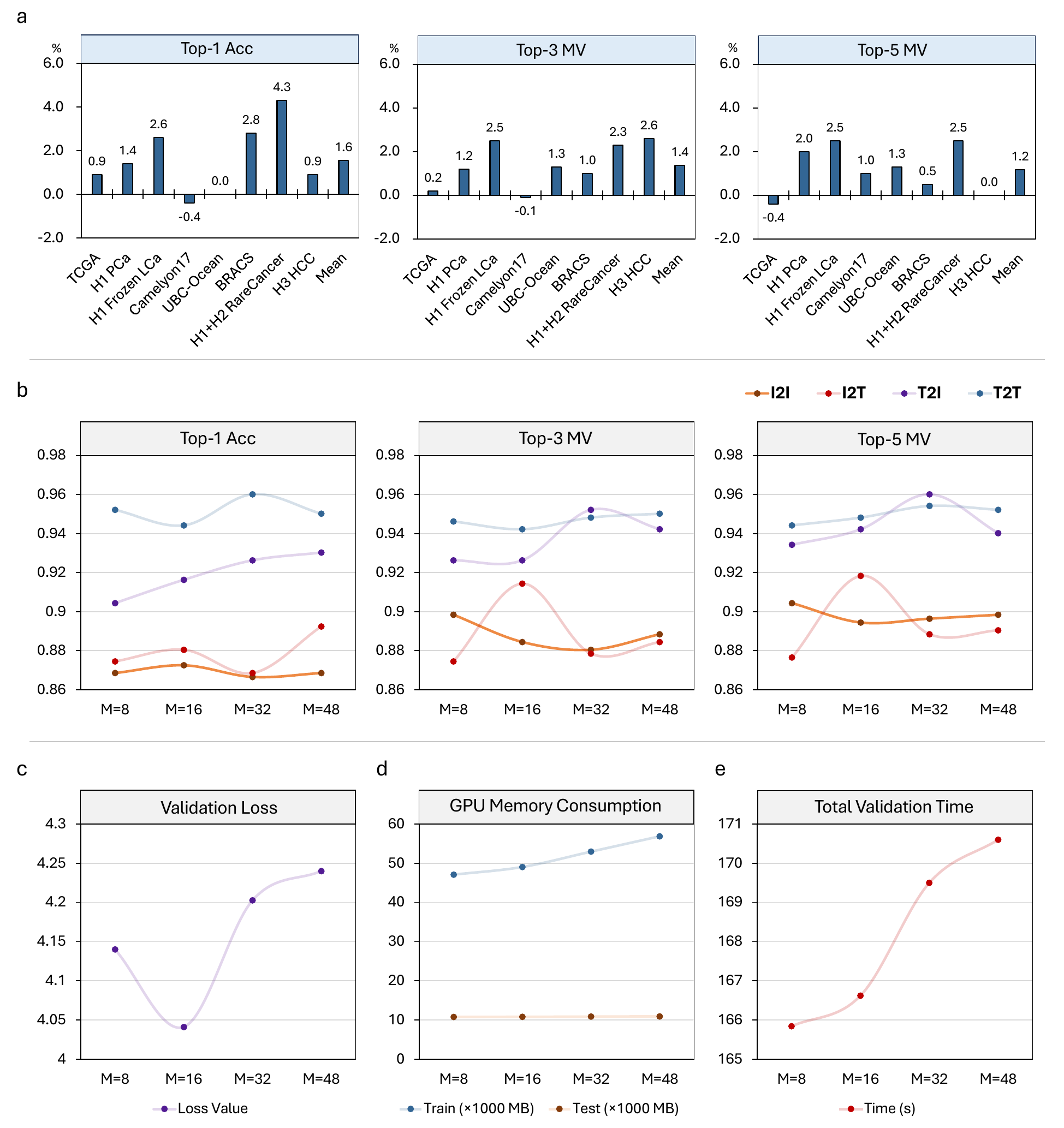}
\caption{\textbf{Ablation analysis of the attentive mosaic mechanism and mosaic-token number $M$.}
\textbf{(a)} Relative change in Acc@1, MV@3, and MV@5 between PathSearch and PathSearch w/o mosaic across eight evaluation tasks: TCGA, H1 PCa, H1 Frozen LCa, Camelyon17, UBC-OCEAN, BRACS, H1+H2 RareCancer, and H3 HCC. Relative change is calculated as $100\times(P_{\mathrm{PathSearch}}-P_{\mathrm{w/o}})/P_{\mathrm{w/o}}$, with the mean across the eight tasks shown for each metric.
\textbf{(b)} Effect of the number of mosaic tokens $M$ on Acc@1, MV@3, and MV@5 on the internal TCGA validation cohort for I2I, I2T, T2I, and T2T retrieval.
\textbf{(c)} Best validation loss obtained for each value of $M$, with the minimum observed at $M=16$.
\textbf{(d,e)} GPU memory consumption and total validation time as a function of $M$.}
\label{parameter_study}
\end{figure*}


\subsection{Ablation Study of PathSearch}

\subsubsection{Contribution of the Attentive Mosaic to Image-to-Image Retrieval}

The attentive mosaic is designed to represent each WSI using a fixed number of compact tokens for fine-grained I2I comparison. Its parallel attention branches aggregate information from a variable number of patch representations, enabling slides with different dimensions and tissue distributions to be represented using the same number of tokens. To quantify the contribution of this component, we compared PathSearch with PathSearch w/o mosaic, a variant in which the mosaic-level distance is excluded and retrieval relies only on the slide-level semantic representation.

The ablation results are summarized in Figure~\ref{parameter_study}(a) and Extended Data Table~\ref{tab:mosaic_ablation}. Relative to PathSearch w/o mosaic, PathSearch showed positive changes in 19 of the 24 dataset-metric comparisons across the eight evaluation tasks. The mean relative changes were +1.6\%, +1.4\%, and +1.2\% for Acc@1, MV@3, and MV@5, respectively. The largest relative improvement in Acc@1 was observed for RareCancer (+4.3\%), whereas the largest improvement in MV@3 was observed for H3 HCC (+2.6\%). For MV@5, H1 Frozen LCa and RareCancer each showed a +2.5\% relative improvement. The remaining five comparisons showed minimal changes, ranging from $-0.4\%$ to 0.0\%. Thus, incorporating the attentive mosaic generally improved I2I retrieval, although the magnitude of the effect varied across datasets and retrieval metrics.

\begin{figure*}
\centering
\includegraphics[width=\textwidth]{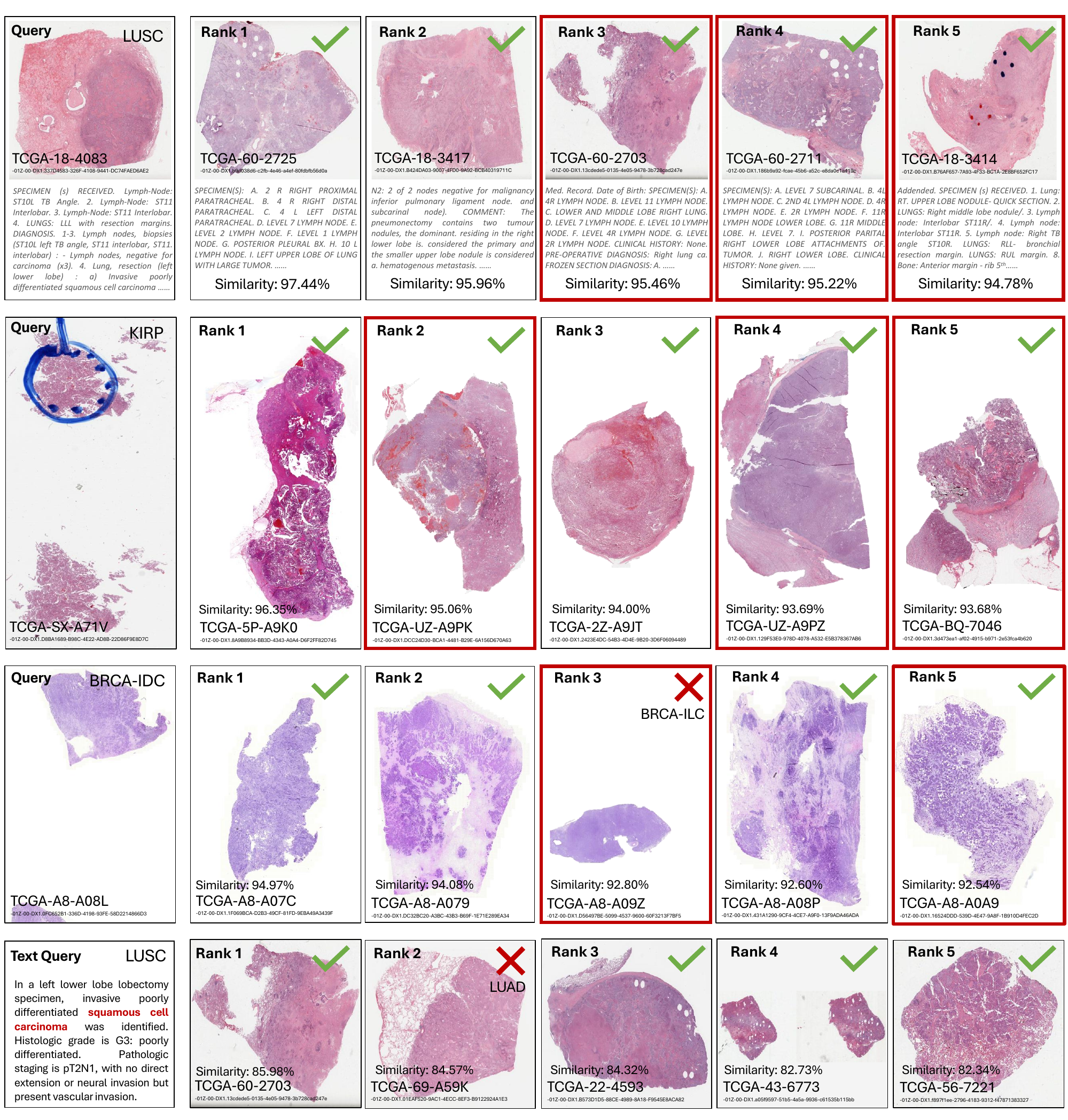}
\caption{\textbf{Selected TCGA retrievals from PathSearch w/o mosaic.} Results are shown for the same queries as Figure~\ref{TCGA_Visualization}; changed retrieved samples are highlighted in \textcolor{red}{red}. \textbf{(Top row)} For the LUSC query, ranks 3-5 differ from the PathSearch results while retaining LUSC labels. \textbf{(Second row)} For the manually marked KIRP query, the ranked samples differ while retaining KIRP labels. \textbf{(Third row)} For the BRCA-IDC query, an ILC slide appears at rank 3 rather than rank 5. \textbf{(Bottom row)} The text-query result is unchanged because the attentive mosaic branch does not contribute to text-based retrieval.}
\label{TCGA_Visualization_wo_bob}
\end{figure*}

\begin{figure*}
\centering
\includegraphics[width=\textwidth]{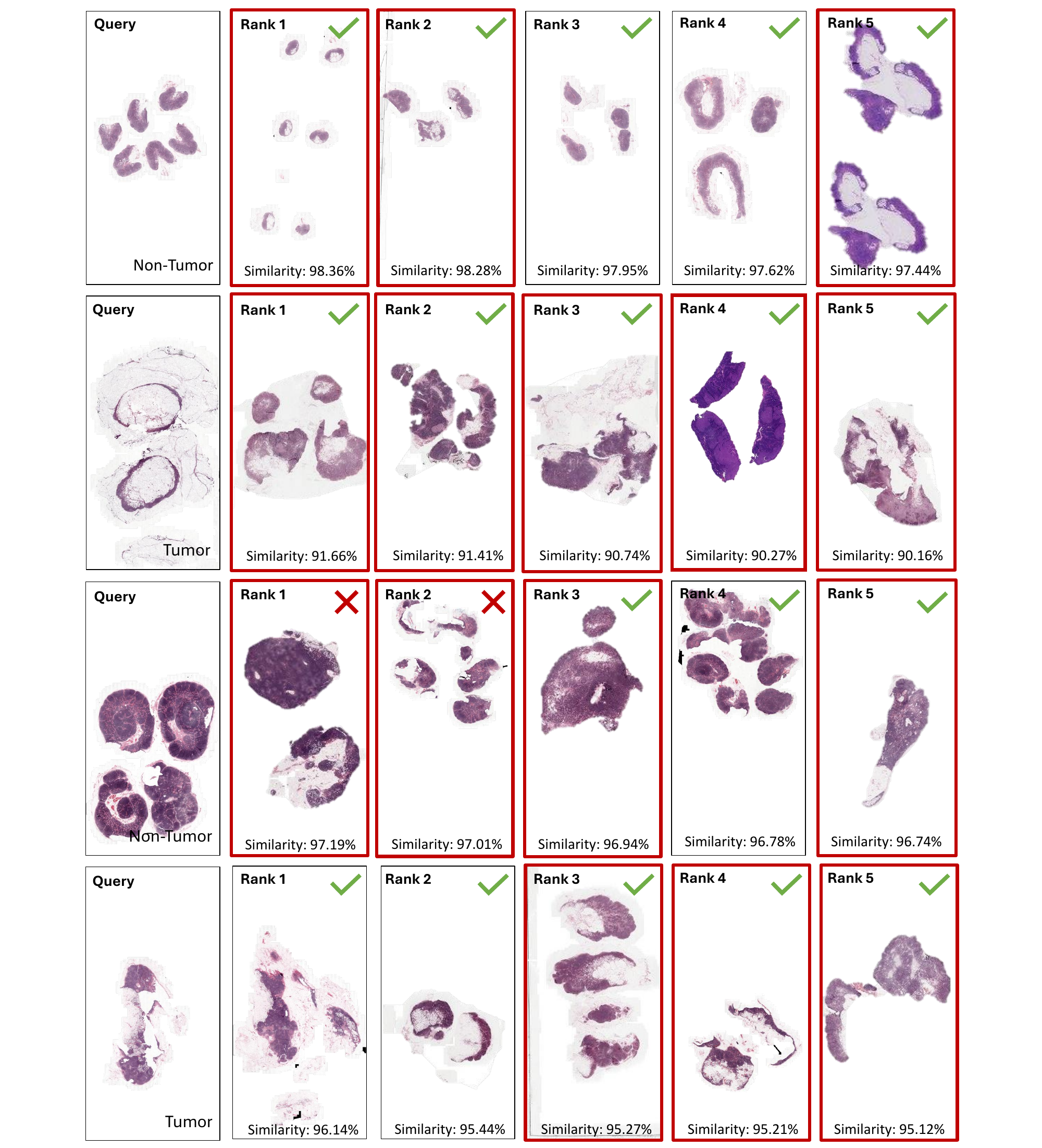}
\caption{\textbf{Selected Camelyon17 retrievals from PathSearch w/o mosaic.} Results are shown for the same four queries as Figure~\ref{Cam17_Visualization}; changed retrieved samples are highlighted in \textcolor{red}{red}. \textbf{(Top row)} Ranks 1, 2, and 5 differ from the PathSearch results. \textbf{(Second row)} All five retrieved samples differ. \textbf{(Third row)} The first two retrieved labels are discordant, changing both Acc@1 and MV@3 for this query. \textbf{(Bottom row)} Ranks 3-5 differ while the first two retrieved samples are unchanged. These selected examples show rank-level changes and should be interpreted together with the aggregate ablation results.}
\label{Cam17_Visualization_wo_bob}
\end{figure*}

Extended Data Figures~\ref{TCGA_Visualization_wo_bob} and~\ref{Cam17_Visualization_wo_bob} provide representative query-level comparisons between PathSearch and PathSearch w/o mosaic. For the displayed BRCA-IDC query, the retrieved ILC slide shifts from rank 5 with PathSearch to rank 3 when the mosaic-level distance is removed. In the displayed Camelyon17 example, removal of the mosaic component alters the labels of the two highest-ranked retrieved slides. These examples illustrate how the attentive mosaic can affect individual retrieval rankings, whereas its overall contribution is quantified by the complete ablation results in Figure~\ref{parameter_study}(a).

\subsubsection{Parameter Study on the Number of Mosaic Tokens}

Because the number of mosaic tokens ($M$) affects representation capacity, computational cost, and storage requirements, we evaluated $M\in\{8,16,32,48\}$ on the 257-case internal TCGA validation cohort (Figure~\ref{parameter_study}(b-e)). For I2I retrieval, Acc@1 was 0.8685, 0.8725, 0.8665, and 0.8685 for $M=8$, 16, 32, and 48, respectively, with the highest value obtained at $M=16$. Performance for the other retrieval directions and majority-vote metrics varied across values of $M$ without a consistent monotonic trend. The minimum validation loss was also observed at $M=16$ (4.0411), compared with 4.1400, 4.2027, and 4.2400 for $M=8$, 32, and 48, respectively.

Increasing $M$ resulted in a corresponding increase in computational and storage requirements. Training GPU memory increased from 47.08\,GB at $M=8$ to 49.04, 52.96, and 56.87\,GB at $M=16$, 32, and 48, respectively, whereas evaluation GPU memory remained relatively stable at 10.78-10.93\,GB. Total validation time increased from 165.84\,s at $M=8$ to 166.62, 169.50, and 170.60\,s at $M=16$, 32, and 48, respectively. The binarized mosaic representations required approximately 1.5\,KB per slide at $M=16$, increasing to approximately 3.0\,KB and 4.5\,KB at $M=32$ and $M=48$, respectively. By comparison, the slide-level semantic representation required approximately 6\,KB per slide.

We therefore selected $M=16$ as the default configuration because it achieved the highest validation I2I Acc@1 and the lowest validation loss while requiring less memory, validation time, and storage than the larger configurations. This choice represents an empirical operating-point trade-off rather than implying that all retrieval metrics are individually optimized at the same value of $M$.

\subsection{Image-to-Image Retrieval Algorithm}

PathSearch combines mosaic-level and semantic-level distances for I2I retrieval, integrating fine-grained morphological comparison with slide-level semantic similarity. As illustrated in Figure~\ref{detailed_pipeline}(c), a query WSI is first tiled and encoded using $E_p$, after which both its attentive mosaic tokens and slide-level semantic representation are generated by $E_m$ and $E_s$.

For mosaic-level comparison, each mosaic token is binarized to enable efficient computation using Hamming distance. Given a query and candidate slide, the Hamming distance is calculated between every pair of query and candidate mosaic tokens. For each query token, the minimum distance to any candidate token is retained, and the median of these per-token minimum distances is used as the slide-level mosaic distance. This procedure enables each query token to match its most similar morphological representation in the candidate slide while reducing sensitivity to individual poorly matched tokens.

For semantic-level comparison, Euclidean distance is calculated between the L2-normalized slide-level embeddings. Because both embeddings have unit norm, Euclidean distance is monotonically related to cosine distance and therefore produces an equivalent ranking when used independently. The mosaic and semantic distance scores are subsequently normalized and combined using a weighted sum,
\begin{equation}
    D_{\mathrm{fuse}}(q,s_i)
    =
    \tilde{D}_{\mathrm{mosaic}}(q,s_i)
    +
    \beta
    \tilde{D}_{\mathrm{sem}}(q,s_i),
\end{equation}
where $\beta$ controls the relative contribution of the semantic representation. Candidate slides are ranked in ascending order of the resulting fused distance. The complete retrieval procedure is provided in Algorithm~\ref{alg:hisra_i2i}, and the mosaic-distance calculation is detailed in Algorithm~\ref{alg:median_min_hamming}.

\subsection{Complexity Analysis of Different Retrieval Paradigms}
\label{extended_complexity_reasoning}

Conventional proportion-based mosaic retrieval methods construct slide representations by retaining a fraction of the available tissue patches. Let $P$ denote the total number of tissue patches in a slide and $R=fP$ the number retained in the mosaic, where $f$ denotes the sampling fraction. For the settings considered in our analysis, $f\in\{0.05,0.10,0.15\}$. For a query slide containing $P_q$ patches and a candidate containing $P_i$ patches, the corresponding mosaic sizes are $R_q=fP_q$ and $R_i=fP_i$. Computing all pairwise distances between the two mosaics therefore requires
\begin{equation}
    O(R_qR_i)
    =
    O(f^2P_qP_i)
\end{equation}
operations per candidate.

Because $f$ is fixed, the asymptotic per-candidate complexity with respect to slide patch count is $O(P_qP_i)$. Assuming an average patch count $\bar{P}$ for both query and candidate slides, this becomes $O(\bar{P}^2)$ per candidate. Exhaustive retrieval over a database containing $S$ slides consequently has a complexity of
\begin{equation}
    O(S\bar{P}^2).
\end{equation}
The quadratic dependence on slide patch count becomes increasingly consequential for large WSIs. For example, when $\bar{P}=5{,}000$ and $f=0.15$, each mosaic contains 750 sampled patches, requiring $750^2=562{,}500$ pairwise token-distance computations for each query-candidate comparison.

PathSearch instead represents every WSI using a fixed number $M$ of attentive mosaic tokens, irrespective of the original patch count. Mosaic-level comparison therefore requires
\begin{equation}
    O(M^2)
\end{equation}
token-distance calculations per candidate. With the default setting $M=16$, this corresponds to 256 token-pair comparisons. Because both $M$ and the dimensionality of each binarized token are fixed, the mosaic-comparison cost is constant with respect to the number of patches in the original WSI. Exhaustive mosaic retrieval across $S$ candidate slides therefore scales as $O(S)$.

PathSearch additionally computes a semantic-level distance between fixed-dimensional slide representations. If the semantic embedding dimension is denoted by $C_s$, the corresponding per-candidate complexity is $O(C_s)$ and the database-level complexity is $O(SC_s)$. Because $C_s$ is fixed, this component also scales linearly with database size, that is, $O(S)$.

The total exhaustive-search complexity of PathSearch can therefore be expressed as
\begin{equation}
    O(SM^2) + O(SC_s).
\end{equation}
With fixed $M$ and $C_s$, this reduces to $O(S)$. PathSearch thus has the same asymptotic scaling with database size as retrieval methods based on a single fixed-dimensional slide embedding, while incurring an additional constant-factor cost for fine-grained mosaic comparison.

\subsection{Statistical Analysis}

Statistical comparisons of retrieval performance were conducted at the query level based on nDCG@10. For each dataset annotated in Figure~\ref{overall_results}, paired outcomes from PathSearch and the best-performing baseline were compared using a two-tailed paired t-test. Accordingly, the reported $P$ annotations in Figure~\ref{overall_results} refer specifically to these pairwise comparisons. For the 95\% confidence intervals presented in the tables, they were estimated using bootstrap resampling with 1,000 iterations. All statistical analyses were performed in Python using SciPy version 1.17.1 \cite{scipy}.

\section{Ethics Approval}

The retrospective analysis of archival pathology slides was conducted in accordance with the Declaration of Helsinki.
This project has been reviewed and approved by the Human and Artifacts Research Ethics Committee (HAREC) of Hong Kong University of Science and Technology. The protocol number is HREP-2025-0459.

\section{Data Availability}

The TCGA data and corresponding labels are available from the National Institutes of Health (NIH) Genomic Data Commons (\url{https://portal.gdc.cancer.gov}). The Camelyon17 dataset is available from the Grand Challenge platform (\url{https://camelyon17.grand-challenge.org/}). Other public evaluation datasets are available from their respective official repositories. The raw in-house H1-H4 datasets are not publicly available due to data-privacy protocol restrictions. 

\section{Code Availability}

The complete source code for PathSearch is publicly available at: \url{https://github.com/Dootmaan/PathSearch}, provided with detailed usage instructions and a small demonstration dataset for rapid verification. PathSearch is released under the GPLv3 License. The implementation incorporates code from CLIP \cite{clip} (\url{https://github.com/openai/CLIP}), developed by OpenAI and released under the MIT License, as well as TransMIL \cite{shao2021transmil} (\url{https://github.com/szc19990412/TransMIL}), released under the GPLv3 License.

\section{Author Contributions}

H.W. and H.C. conceived the idea and designed the work. H.W. and Zhengjie Z. constructed the PathSearch workflow and conducted the model training. For downstream tasks, H.W. was responsible for PathSearch and the other mosaic-based methods across the evaluated datasets, including the HCC-related experiments; Zhengjie Z. and J.H. was responsible for experiments with the other methods on the public datasets; and J.M. contributed to experiments reported in Extended Data. F.W., Y.S., Q.C., J.W., and Zhengyu Z. curated the in-house datasets and helped organize the reader study with B.L., X.Z., and Li L. YW.C. and Lanfen L. provided medical guidance. H.W. and Zhengjie Z. wrote the manuscript with input from all authors. All authors reviewed and approved the final manuscript. H.C. supervised the research.

\section{Funding}
This work was supported by the National Natural Science Foundation of China (No. 62202403), the National Key Research and Development Program of China (No. 2022YFC2504605 and No. 2023YFE0204000), the Innovation and Technology Commission of Hong Kong (Project No. MHP/002/22 and ITCPD/17-9), and the Research Grants Council of the Hong Kong Special Administrative Region, China (Project No. T45-401/22-N). This work was also supported in part by the Grant-in-Aid for Scientific Research from the Ministry of Education, Culture, Sports, Science and Technology (MEXT), Japan (No. 20KK0234 and No. 21H03470), by JST CREST (No. JPMJCR25T4), and by the JST Sakura Science Program, Japan.

\section{Acknowledgments}
The results presented here are in part derived from data of the TCGA Research Network (https://www.cancer.gov/tcga). The authors gratefully acknowledge the contributions of the patients, the specimen donors, and the research groups associated with each TCGA Project.

The authors also would like to thank Yunjun Yang from H3 for curating the HCC risk evaluation dataset, and Ruiqi Yan from Zhejiang University for assistance with the reproduction of baseline methods during experimentation.

\section{Competing Interests}
The authors declare no competing interests.

\bibliography{sample.bib}

\section*{Extended Data}\label{Extended Data}
\clearpage

\newcommand{\best}[1]{\fontseries{b}\selectfont #1\normalfont}

\subsection*{Detailed Results for the Benchmark}

Tables~\ref{tab:tcga_img2img}-\ref{tab:fahwmu_hcc_risk} report the exact task-level results underlying the benchmark figure, including multimodal retrieval on TCGA and H1 PCa and image-to-image retrieval on the remaining datasets.

\begin{table}[t]
\caption{\textbf{Image-to-image (I2I) retrieval performance comparison on the TCGA test set.} The best and second-best results are highlighted in \textbf{bold} and \underline{underline}, respectively. Values are percentages; bracketed ranges are 95\% confidence intervals.}
\label{tab:tcga_img2img}
\centering
\begin{tabular}{@{}lccc@{}}
\toprule
Method            & Acc@1 & MV@3 & MV@5 \\ \midrule
Yottixel & \shortstack{85.0\\{\scriptsize [81.9--88.0]}} & \shortstack{\underline{87.5}\\{\scriptsize [84.5--90.4]}} & \shortstack{\underline{89.6}\\{\scriptsize [86.9--92.0]}} \\
SISH & \shortstack{\underline{87.3}\\{\scriptsize [84.3--90.0]}} & \shortstack{\underline{87.5}\\{\scriptsize [84.7--90.2]}} & \shortstack{87.5\\{\scriptsize [84.5--90.2]}} \\
RetCCL & \shortstack{81.7\\{\scriptsize [78.3--85.1]}} & \shortstack{82.3\\{\scriptsize [78.9--85.7]}} & \shortstack{81.9\\{\scriptsize [78.5--85.3]}} \\
GigaPath & \shortstack{66.3\\{\scriptsize [62.4--70.3]}} & \shortstack{70.9\\{\scriptsize [66.7--74.7]}} & \shortstack{70.1\\{\scriptsize [66.1--74.1]}} \\
CHIEF & \shortstack{77.1\\{\scriptsize [73.5--80.7]}} & \shortstack{75.1\\{\scriptsize [71.3--78.9]}} & \shortstack{78.1\\{\scriptsize [74.3--81.7]}} \\ \midrule
FGCR & \shortstack{48.0\\{\scriptsize [43.8--52.4]}} & \shortstack{47.8\\{\scriptsize [43.4--52.0]}} & \shortstack{50.8\\{\scriptsize [46.4--55.2]}} \\
CONCH & \shortstack{50.4\\{\scriptsize [46.0--54.8]}} & \shortstack{63.9\\{\scriptsize [59.8--68.1]}} & \shortstack{60.0\\{\scriptsize [55.6--63.9]}} \\ \midrule
PathSearch & \shortstack{\textbf{91.4}\\{\scriptsize [89.0--93.8]}} & \shortstack{\textbf{91.6}\\{\scriptsize [89.2--94.0]}} & \shortstack{\textbf{92.0}\\{\scriptsize [89.6--94.2]}} \\ \bottomrule
\end{tabular}
\end{table}

\begin{table*}[t]
\caption{\textbf{Cross-modal and text-to-text retrieval performance on the TCGA test set.} Mean denotes the average Acc@1 across I2I, I2T, T2I, and T2T. The best and second-best results are in \textbf{bold} and \underline{underline}. Values are percentages; bracketed ranges are 95\% confidence intervals.}
\label{tab:tcga_crossmodal}
\centering
\resizebox{\textwidth}{!}{%
\begin{tabular}{@{}lccccccccc|c@{}}
\toprule
\multirow{2}{*}{Method} & \multicolumn{3}{c}{I2T} & \multicolumn{3}{c}{T2I} & \multicolumn{3}{c}{T2T} & \multirow{2}{*}{Mean} \\
\cmidrule(lr){2-4} \cmidrule(lr){5-7} \cmidrule(lr){8-10}
& Acc@1 & MV@3 & MV@5 & Acc@1 & MV@3 & MV@5 & Acc@1 & MV@3 & MV@5 & \\ \midrule
FGCR & \shortstack{54.0\\{\scriptsize [49.8--58.6]}} & \shortstack{45.8\\{\scriptsize [41.6--50.2]}} & \shortstack{45.8\\{\scriptsize [41.4--50.2]}} & \shortstack{\underline{45.8}\\{\scriptsize [41.4--50.4]}} & \shortstack{\underline{45.8}\\{\scriptsize [41.4--50.2]}} & \shortstack{\underline{45.8}\\{\scriptsize [41.4--50.4]}} & \shortstack{\underline{62.0}\\{\scriptsize [57.8--66.3]}} & \shortstack{\underline{63.9}\\{\scriptsize [59.8--67.9]}} & \shortstack{\underline{65.1}\\{\scriptsize [61.0--69.3]}} & \shortstack{\underline{52.5}\\{\scriptsize [48.2--56.8]}} \\
CONCH & \shortstack{\underline{90.2}\\{\scriptsize [87.6--92.6]}} & \shortstack{\underline{90.8}\\{\scriptsize [88.2--93.2]}} & \shortstack{\underline{90.8}\\{\scriptsize [88.2--93.2]}} & \shortstack{20.3\\{\scriptsize [16.7--23.9]}} & \shortstack{23.7\\{\scriptsize [19.9--27.5]}} & \shortstack{26.9\\{\scriptsize [22.9--30.9]}} & \shortstack{32.7\\{\scriptsize [28.7--36.7]}} & \shortstack{29.7\\{\scriptsize [25.7--33.7]}} & \shortstack{30.1\\{\scriptsize [26.1--34.1]}} & \shortstack{48.4\\{\scriptsize [44.2--52.8]}} \\ \midrule
PathSearch & \shortstack{\textbf{95.0}\\{\scriptsize [93.2--96.8]}} & \shortstack{\textbf{94.6}\\{\scriptsize [92.6--96.4]}} & \shortstack{\textbf{95.6}\\{\scriptsize [93.8--97.2]}} & \shortstack{\textbf{88.8}\\{\scriptsize [86.1--91.6]}} & \shortstack{\textbf{89.4}\\{\scriptsize [86.5--92.0]}} & \shortstack{\textbf{88.7}\\{\scriptsize [86.0--91.3]}} & \shortstack{\textbf{92.4}\\{\scriptsize [90.0--94.8]}} & \shortstack{\textbf{95.8}\\{\scriptsize [93.8--97.6]}} & \shortstack{\textbf{95.2}\\{\scriptsize [93.4--96.8]}} & \shortstack{\textbf{91.9}\\{\scriptsize [89.5--94.3]}} \\ \bottomrule
\end{tabular}
}
\end{table*}

\begin{table}[t]
\caption{\textbf{I2I retrieval performance comparison on the H1 PCa test set.} The best and second-best results are highlighted in \textbf{bold} and \underline{underline}, respectively. Values are percentages; bracketed ranges are 95\% confidence intervals.}
\label{tab:h1_pca_img2img}
\centering
\begin{tabular}{@{}lccc@{}}
\toprule

Method           & Acc@1         & MV@3          & MV@5          \\ \midrule
Yottixel & \shortstack{\underline{82.7}\\{\scriptsize [79.6--85.8]}} & \shortstack{\underline{84.3}\\{\scriptsize [81.3--87.2]}} & \shortstack{\underline{84.2}\\{\scriptsize [81.3--87.0]}} \\
SISH & \shortstack{82.3\\{\scriptsize [79.2--85.3]}} & \shortstack{84.1\\{\scriptsize [81.1--87.0]}} & \shortstack{\underline{84.2}\\{\scriptsize [81.1--87.2]}} \\
RetCCL & \shortstack{82.2\\{\scriptsize [79.1--85.3]}} & \shortstack{83.6\\{\scriptsize [80.6--86.7]}} & \shortstack{83.1\\{\scriptsize [79.9--86.2]}} \\
GigaPath & \shortstack{81.3\\{\scriptsize [78.0--84.4]}} & \shortstack{82.4\\{\scriptsize [79.2--85.5]}} & \shortstack{81.5\\{\scriptsize [78.4--84.6]}} \\
CHIEF & \shortstack{81.7\\{\scriptsize [78.4--84.8]}} & \shortstack{82.9\\{\scriptsize [79.9--86.0]}} & \shortstack{82.2\\{\scriptsize [79.1--85.5]}} \\ \midrule
FGCR & \shortstack{67.5\\{\scriptsize [63.7--71.1]}} & \shortstack{68.0\\{\scriptsize [64.4--72.0]}} & \shortstack{67.2\\{\scriptsize [63.5--70.9]}} \\
CONCH & \shortstack{81.2\\{\scriptsize [78.0--84.3]}} & \shortstack{82.1\\{\scriptsize [78.9--85.1]}} & \shortstack{81.2\\{\scriptsize [78.0--84.3]}} \\ \midrule
PathSearch & \shortstack{\textbf{83.9}\\{\scriptsize [80.8--86.9]}} & \shortstack{\textbf{85.8}\\{\scriptsize [82.9--88.6]}} & \shortstack{\textbf{86.2}\\{\scriptsize [83.4--89.0]}} \\ \bottomrule
\end{tabular}

\end{table}
\begin{table*}[t]
\caption{\textbf{Multimodal retrieval performance on the H1 PCa benign-versus-malignant task.} I2T, T2I, and T2T denote image-to-text, text-to-image, and text-to-text retrieval, respectively. Mean denotes the average Acc@1 across I2I, I2T, T2I, and T2T. The best and second-best results are highlighted in \textbf{bold} and \underline{underline}. Values are percentages; bracketed ranges are 95\% confidence intervals.}
\label{tab:h1_pca_multimodal}
\centering
\resizebox{\textwidth}{!}{%
\begin{tabular}{@{}lcccccccccc@{}}
\toprule
\multirow{2}{*}{Method} & \multicolumn{3}{c}{I2T} & \multicolumn{3}{c}{T2I} & \multicolumn{3}{c}{T2T} & \multirow{2}{*}{Mean} \\
\cmidrule(lr){2-4} \cmidrule(lr){5-7} \cmidrule(lr){8-10}
& Acc@1 & MV@3 & MV@5 & Acc@1 & MV@3 & MV@5 & Acc@1 & MV@3 & MV@5 & \\ \midrule
FGCR & \shortstack{60.1\\{\scriptsize [56.1--64.0]}} & \shortstack{64.7\\{\scriptsize [60.9--68.7]}} & \shortstack{64.9\\{\scriptsize [61.1--68.7]}} & \shortstack{60.6\\{\scriptsize [56.4--64.5]}} & \shortstack{65.3\\{\scriptsize [61.4--69.2]}} & \shortstack{65.6\\{\scriptsize [61.6--69.4]}} & \shortstack{65.9\\{\scriptsize [61.9--69.7]}} & \shortstack{66.3\\{\scriptsize [62.5--70.1]}} & \shortstack{63.7\\{\scriptsize [59.7--67.5]}} & \shortstack{63.5\\{\scriptsize [61.6--65.5]}} \\
CONCH & \shortstack{\underline{75.0}\\{\scriptsize [71.5--78.5]}} & \shortstack{\underline{74.8}\\{\scriptsize [71.3--78.4]}} & \shortstack{\underline{74.8}\\{\scriptsize [71.3--78.2]}} & \shortstack{\underline{75.0}\\{\scriptsize [71.5--78.5]}} & \shortstack{\underline{79.8}\\{\scriptsize [76.6--82.9]}} & \shortstack{\underline{79.8}\\{\scriptsize [76.3--83.0]}} & \shortstack{\underline{71.2}\\{\scriptsize [67.5--75.1]}} & \shortstack{\underline{71.5}\\{\scriptsize [67.8--75.1]}} & \shortstack{\underline{79.6}\\{\scriptsize [76.5--82.9]}} & \shortstack{\underline{75.6}\\{\scriptsize [73.8--77.3]}} \\ \midrule
PathSearch & \shortstack{\textbf{79.8}\\{\scriptsize [76.5--83.0]}} & \shortstack{\textbf{79.8}\\{\scriptsize [76.5--83.0]}} & \shortstack{\textbf{79.6}\\{\scriptsize [76.3--82.8]}} & \shortstack{\textbf{81.2}\\{\scriptsize [77.9--84.3]}} & \shortstack{\textbf{81.5}\\{\scriptsize [78.2--84.6]}} & \shortstack{\textbf{82.0}\\{\scriptsize [78.7--85.1]}} & \shortstack{\textbf{99.8}\\{\scriptsize [99.3--100.0]}} & \shortstack{\textbf{99.8}\\{\scriptsize [99.3--100.0]}} & \shortstack{\textbf{99.8}\\{\scriptsize [99.3--100.0]}} & \shortstack{\textbf{86.2}\\{\scriptsize [84.8--87.5]}} \\ \bottomrule
\end{tabular}
}
\end{table*}

\begin{table}[t]
\caption{\textbf{Performance on the multi-center Camelyon17 dataset for metastasis detection.} The best and second-best results are highlighted in \textbf{bold} and \underline{underline}. Values are percentages; bracketed ranges are 95\% confidence intervals.}
\label{tab:camelyon17}
\centering
\begin{tabular}{@{}lcccc@{}}
\toprule
Method            & Acc@1 & MV@3 & MV@5 & MV@10 \\ \midrule
Yottixel & \shortstack{77.9\\{\scriptsize [75.3--80.5]}} & \shortstack{78.0\\{\scriptsize [75.3--80.7]}} & \shortstack{78.4\\{\scriptsize [75.7--81.0]}} & \shortstack{\underline{79.1}\\{\scriptsize [76.5--81.7]}} \\
SISH & \shortstack{78.0\\{\scriptsize [75.3--80.7]}} & \shortstack{78.0\\{\scriptsize [75.3--80.6]}} & \shortstack{\underline{78.8}\\{\scriptsize [76.1--81.4]}} & \shortstack{78.8\\{\scriptsize [76.1--81.5]}} \\
RetCCL & \shortstack{72.4\\{\scriptsize [69.5--75.3]}} & \shortstack{72.5\\{\scriptsize [69.6--75.4]}} & \shortstack{72.4\\{\scriptsize [69.5--75.4]}} & \shortstack{73.1\\{\scriptsize [70.3--75.9]}} \\
GigaPath & \shortstack{62.7\\{\scriptsize [59.5--65.8]}} & \shortstack{63.3\\{\scriptsize [60.2--66.4]}} & \shortstack{64.4\\{\scriptsize [61.2--67.5]}} & \shortstack{64.4\\{\scriptsize [61.1--67.4]}} \\
CHIEF & \shortstack{\underline{78.7}\\{\scriptsize [76.1--81.3]}} & \shortstack{\underline{78.8}\\{\scriptsize [76.2--81.5]}} & \shortstack{78.4\\{\scriptsize [75.6--81.0]}} & \shortstack{78.2\\{\scriptsize [75.3--80.8]}} \\ \midrule
FGCR & \shortstack{57.9\\{\scriptsize [54.7--60.9]}} & \shortstack{58.0\\{\scriptsize [54.8--61.1]}} & \shortstack{59.2\\{\scriptsize [56.0--62.5]}} & \shortstack{60.1\\{\scriptsize [56.8--63.3]}} \\
CONCH & \shortstack{62.6\\{\scriptsize [59.5--65.7]}} & \shortstack{61.7\\{\scriptsize [58.6--64.8]}} & \shortstack{64.3\\{\scriptsize [61.1--67.4]}} & \shortstack{63.9\\{\scriptsize [60.6--67.0]}} \\ \midrule
PathSearch & \shortstack{\textbf{81.3}\\{\scriptsize [78.7--83.9]}} & \shortstack{\textbf{83.9}\\{\scriptsize [81.5--86.1]}} & \shortstack{\textbf{84.1}\\{\scriptsize [81.6--86.6]}} & \shortstack{\textbf{83.4}\\{\scriptsize [81.0--85.7]}} \\ \bottomrule
\end{tabular}

\end{table}

\begin{table}[t]
\caption{\textbf{Performance comparison for malignancy detection on the in-house frozen lung cancer dataset.} The best and second-best results are highlighted in \textbf{bold} and \underline{underline}. Values are percentages; bracketed ranges are 95\% confidence intervals.}
\label{tab:frozen_lca}
\centering
\begin{tabular}{@{}lcccc@{}}
\toprule
Method            & Acc@1 & MV@3 & MV@5 & MV@10 \\ \midrule
Yottixel & \shortstack{80.3\\{\scriptsize [77.0--83.5]}} & \shortstack{\underline{82.6}\\{\scriptsize [79.5--85.6]}} & \shortstack{82.8\\{\scriptsize [79.5--85.8]}} & \shortstack{\underline{84.0}\\{\scriptsize [81.0--87.0]}} \\
SISH & \shortstack{\underline{80.7}\\{\scriptsize [77.4--83.8]}} & \shortstack{82.3\\{\scriptsize [79.2--85.4]}} & \shortstack{\underline{83.2}\\{\scriptsize [80.1--86.3]}} & \shortstack{83.6\\{\scriptsize [80.4--86.7]}} \\
RetCCL & \shortstack{76.9\\{\scriptsize [73.5--80.2]}} & \shortstack{78.4\\{\scriptsize [75.1--81.7]}} & \shortstack{79.2\\{\scriptsize [75.8--82.4]}} & \shortstack{79.6\\{\scriptsize [76.3--82.9]}} \\
GigaPath & \shortstack{68.8\\{\scriptsize [64.9--72.4]}} & \shortstack{71.3\\{\scriptsize [67.4--75.1]}} & \shortstack{71.8\\{\scriptsize [68.1--75.4]}} & \shortstack{73.3\\{\scriptsize [69.8--77.0]}} \\
CHIEF & \shortstack{74.4\\{\scriptsize [70.8--78.1]}} & \shortstack{76.7\\{\scriptsize [73.1--80.1]}} & \shortstack{76.5\\{\scriptsize [73.0--79.9]}} & \shortstack{75.8\\{\scriptsize [72.2--79.4]}} \\ \midrule
FGCR & \shortstack{72.1\\{\scriptsize [68.3--75.8]}} & \shortstack{74.1\\{\scriptsize [70.5--77.6]}} & \shortstack{73.7\\{\scriptsize [70.1--77.2]}} & \shortstack{74.4\\{\scriptsize [70.8--77.9]}} \\
CONCH & \shortstack{79.0\\{\scriptsize [75.6--82.6]}} & \shortstack{80.1\\{\scriptsize [76.7--83.3]}} & \shortstack{79.9\\{\scriptsize [76.5--83.1]}} & \shortstack{80.4\\{\scriptsize [77.0--83.6]}} \\ \midrule
PathSearch & \shortstack{\textbf{82.3}\\{\scriptsize [78.9--85.5]}} & \shortstack{\textbf{85.0}\\{\scriptsize [81.8--88.0]}} & \shortstack{\textbf{85.8}\\{\scriptsize [82.7--88.4]}} & \shortstack{\textbf{86.6}\\{\scriptsize [83.4--89.5]}} \\ \bottomrule
\end{tabular}

\end{table}

\begin{table}[t]
\caption{\textbf{Retrieval results for ovarian cancer subtype classification on the UBC-OCEAN dataset.} The best and second-best results are highlighted in \textbf{bold} and \underline{underline}. Values are percentages; bracketed ranges are 95\% confidence intervals.}
\label{tab:ubc_ocean}
\centering
\begin{tabular}{@{}lcccc@{}}
\toprule
Method            & Acc@1 & MV@3 & MV@5 & MV@10 \\ \midrule
Yottixel & \shortstack{\underline{77.7}\\{\scriptsize [73.9--81.3]}} & \shortstack{\underline{80.1}\\{\scriptsize [76.6--83.4]}} & \shortstack{80.5\\{\scriptsize [77.2--84.0]}} & \shortstack{\underline{81.0}\\{\scriptsize [77.6--84.4]}} \\
SISH & \shortstack{73.8\\{\scriptsize [70.0--77.6]}} & \shortstack{79.9\\{\scriptsize [76.4--83.2]}} & \shortstack{\underline{81.0}\\{\scriptsize [77.6--84.2]}} & \shortstack{79.9\\{\scriptsize [76.2--83.4]}} \\
RetCCL & \shortstack{75.4\\{\scriptsize [71.5--78.9]}} & \shortstack{79.2\\{\scriptsize [75.6--82.7]}} & \shortstack{79.7\\{\scriptsize [76.0--83.0]}} & \shortstack{80.3\\{\scriptsize [76.8--83.6]}} \\
GigaPath & \shortstack{77.1\\{\scriptsize [73.5--80.7]}} & \shortstack{78.4\\{\scriptsize [74.9--81.9]}} & \shortstack{79.9\\{\scriptsize [76.4--83.2]}} & \shortstack{79.9\\{\scriptsize [76.4--83.0]}} \\
CHIEF & \shortstack{\textbf{79.4}\\{\scriptsize [75.8--82.8]}} & \shortstack{\underline{80.1}\\{\scriptsize [76.6--83.4]}} & \shortstack{80.9\\{\scriptsize [77.4--84.2]}} & \shortstack{79.9\\{\scriptsize [76.4--83.4]}} \\ \midrule
FGCR & \shortstack{73.1\\{\scriptsize [69.2--76.8]}} & \shortstack{75.7\\{\scriptsize [71.9--79.3]}} & \shortstack{76.2\\{\scriptsize [72.5--79.9]}} & \shortstack{76.0\\{\scriptsize [72.3--79.7]}} \\
CONCH & \shortstack{76.9\\{\scriptsize [73.2--80.6]}} & \shortstack{78.1\\{\scriptsize [74.3--81.8]}} & \shortstack{80.7\\{\scriptsize [77.0--84.0]}} & \shortstack{79.9\\{\scriptsize [76.4--83.2]}} \\ \midrule
PathSearch & \shortstack{75.8\\{\scriptsize [72.1--79.5]}} & \shortstack{\textbf{80.5}\\{\scriptsize [77.2--83.6]}} & \shortstack{\textbf{82.7}\\{\scriptsize [79.3--86.0]}} & \shortstack{\textbf{82.9}\\{\scriptsize [79.6--86.1]}} \\ \bottomrule
\end{tabular}%

\end{table}

\begin{table}[t]
\caption{\textbf{Retrieval performance on the BRACS Atypical-versus-Malignant task.} The best and second-best results are highlighted in \textbf{bold} and \underline{underline}. Values are percentages; bracketed ranges are 95\% confidence intervals.}
\label{tab:bracs}
\centering
\begin{tabular}{@{}lcccc@{}}
\toprule
Method & Acc@1 & MV@3 & MV@5 & MV@10 \\ \midrule
Yottixel & \shortstack{73.9\\{\scriptsize [68.8--79.1]}} & \shortstack{76.6\\{\scriptsize [71.6--81.6]}} & \shortstack{76.9\\{\scriptsize [72.0--81.9]}} & \shortstack{77.2\\{\scriptsize [72.3--81.9]}} \\
SISH & \shortstack{73.6\\{\scriptsize [68.4--78.7]}} & \shortstack{77.2\\{\scriptsize [72.3--81.9]}} & \shortstack{76.6\\{\scriptsize [71.6--81.6]}} & \shortstack{77.9\\{\scriptsize [73.0--82.6]}} \\
RetCCL & \shortstack{71.9\\{\scriptsize [66.7--77.0]}} & \shortstack{75.4\\{\scriptsize [70.2--80.5]}} & \shortstack{75.4\\{\scriptsize [70.2--80.5]}} & \shortstack{75.8\\{\scriptsize [70.6--80.9]}} \\
GigaPath & \shortstack{71.6\\{\scriptsize [66.3--76.6]}} & \shortstack{69.2\\{\scriptsize [63.8--74.5]}} & \shortstack{72.7\\{\scriptsize [67.4--77.7]}} & \shortstack{70.2\\{\scriptsize [64.9--75.2]}} \\
CHIEF & \shortstack{\underline{74.8}\\{\scriptsize [69.9--79.8]}} & \shortstack{\underline{78.7}\\{\scriptsize [73.8--83.3]}} & \shortstack{\underline{79.1}\\{\scriptsize [74.1--84.0]}} & \shortstack{\underline{80.5}\\{\scriptsize [75.9--85.1]}} \\ \midrule
FGCR & \shortstack{70.9\\{\scriptsize [65.6--76.2]}} & \shortstack{72.3\\{\scriptsize [66.7--77.3]}} & \shortstack{71.6\\{\scriptsize [66.0--76.6]}} & \shortstack{72.0\\{\scriptsize [66.7--77.3]}} \\
CONCH & \shortstack{73.8\\{\scriptsize [68.8--78.7]}} & \shortstack{75.2\\{\scriptsize [70.2--80.5]}} & \shortstack{73.4\\{\scriptsize [68.1--78.4]}} & \shortstack{75.2\\{\scriptsize [70.2--80.1]}} \\ \midrule
PathSearch & \shortstack{\textbf{75.9}\\{\scriptsize [70.6--80.9]}} & \shortstack{\textbf{80.1}\\{\scriptsize [74.8--84.8]}} & \shortstack{\textbf{81.2}\\{\scriptsize [76.2--85.8]}} & \shortstack{\textbf{81.2}\\{\scriptsize [76.6--85.5]}} \\ \bottomrule
\end{tabular}%

\end{table}

\begin{table}[t]
\caption{\textbf{Retrieval performance on the H1+H2 RareCancer mixed-gallery task.} The best and second-best results are highlighted in \textbf{bold} and \underline{underline}. Values are percentages; bracketed ranges are 95\% confidence intervals.}
\label{tab:rare_cancer}
\centering
\begin{tabular}{@{}lcccc@{}}
\toprule
Method & Acc@1 & MV@3 & MV@5 & MV@10 \\ \midrule
Yottixel & \shortstack{95.5\\{\scriptsize [92.4--98.0]}} & \shortstack{96.0\\{\scriptsize [92.9--98.5]}} & \shortstack{\underline{96.0}\\{\scriptsize [92.9--98.5]}} & \shortstack{96.0\\{\scriptsize [93.4--98.5]}} \\
SISH & \shortstack{\underline{96.5}\\{\scriptsize [93.9--99.0]}} & \shortstack{\underline{96.5}\\{\scriptsize [93.9--99.0]}} & \shortstack{\underline{96.0}\\{\scriptsize [92.9--98.5]}} & \shortstack{96.0\\{\scriptsize [92.9--98.5]}} \\
RetCCL & \shortstack{89.9\\{\scriptsize [85.4--93.9]}} & \shortstack{90.4\\{\scriptsize [86.4--94.4]}} & \shortstack{89.9\\{\scriptsize [85.4--93.9]}} & \shortstack{89.4\\{\scriptsize [84.8--93.4]}} \\
GigaPath & \shortstack{90.4\\{\scriptsize [85.9--94.4]}} & \shortstack{90.0\\{\scriptsize [85.9--93.9]}} & \shortstack{89.5\\{\scriptsize [84.8--93.4]}} & \shortstack{89.0\\{\scriptsize [84.3--92.9]}} \\
CHIEF & \shortstack{88.9\\{\scriptsize [84.3--92.9]}} & \shortstack{89.4\\{\scriptsize [84.8--93.4]}} & \shortstack{88.9\\{\scriptsize [84.8--92.9]}} & \shortstack{88.4\\{\scriptsize [83.8--92.4]}} \\ \midrule
FGCR & \shortstack{84.9\\{\scriptsize [79.8--89.9]}} & \shortstack{85.9\\{\scriptsize [80.8--90.4]}} & \shortstack{86.4\\{\scriptsize [81.3--90.9]}} & \shortstack{86.9\\{\scriptsize [81.8--91.4]}} \\
CONCH & \shortstack{94.4\\{\scriptsize [91.4--97.5]}} & \shortstack{95.96\\{\scriptsize [92.9--98.5]}} & \shortstack{95.96\\{\scriptsize [92.9--98.5]}} & \shortstack{\underline{96.46}\\{\scriptsize [93.9--99.0]}} \\ \midrule
PathSearch & \shortstack{\textbf{97.5}\\{\scriptsize [94.9--99.5]}} & \shortstack{\textbf{98.0}\\{\scriptsize [96.0--99.5]}} & \shortstack{\textbf{97.5}\\{\scriptsize [94.9--99.5]}} & \shortstack{\textbf{97.5}\\{\scriptsize [94.9--99.5]}} \\ \bottomrule
\end{tabular}%

\end{table}

\begin{table}[t]
\caption{\textbf{Performance on the binary task of HCC risk evaluation using the in-house H3 HCC dataset.} The best and second-best results are highlighted in \textbf{bold} and \underline{underline}. Values are percentages; bracketed ranges are 95\% confidence intervals.}
\label{tab:fahwmu_hcc_risk}
\centering
\begin{tabular}{@{}lcccc@{}}
\toprule
Method            & Acc@1 & MV@3 & MV@5 & MV@10 \\ \midrule
Yottixel & \shortstack{84.5\\{\scriptsize [79.7--89.2]}} & \shortstack{82.1\\{\scriptsize [77.1--87.0]}} & \shortstack{78.8\\{\scriptsize [73.6--84.0]}} & \shortstack{80.7\\{\scriptsize [75.8--85.7]}} \\
SISH & \shortstack{\underline{84.8}\\{\scriptsize [79.7--89.2]}} & \shortstack{\underline{82.7}\\{\scriptsize [77.9--87.4]}} & \shortstack{79.2\\{\scriptsize [73.6--84.4]}} & \shortstack{\underline{81.1}\\{\scriptsize [75.8--86.1]}} \\
RetCCL & \shortstack{78.8\\{\scriptsize [73.2--84.0]}} & \shortstack{77.9\\{\scriptsize [72.7--83.1]}} & \shortstack{77.9\\{\scriptsize [72.7--83.1]}} & \shortstack{78.2\\{\scriptsize [72.7--83.5]}} \\
GigaPath & \shortstack{63.3\\{\scriptsize [57.1--69.3]}} & \shortstack{64.5\\{\scriptsize [58.4--70.6]}} & \shortstack{64.5\\{\scriptsize [58.4--70.6]}} & \shortstack{64.5\\{\scriptsize [58.4--70.6]}} \\
CHIEF & \shortstack{81.4\\{\scriptsize [76.2--86.7]}} & \shortstack{79.7\\{\scriptsize [74.5--85.0]}} & \shortstack{\underline{80.1}\\{\scriptsize [74.9--84.8]}} & \shortstack{80.7\\{\scriptsize [75.3--85.7]}} \\ \midrule
FGCR & \shortstack{58.4\\{\scriptsize [51.9--64.5]}} & \shortstack{56.7\\{\scriptsize [50.2--62.8]}} & \shortstack{57.1\\{\scriptsize [50.6--63.2]}} & \shortstack{57.6\\{\scriptsize [51.1--64.1]}} \\
CONCH & \shortstack{62.3\\{\scriptsize [56.3--68.4]}} & \shortstack{62.0\\{\scriptsize [55.4--68.0]}} & \shortstack{62.0\\{\scriptsize [55.8--68.4]}} & \shortstack{62.2\\{\scriptsize [55.8--68.4]}} \\ \midrule
PathSearch & \shortstack{\textbf{88.7}\\{\scriptsize [84.4--92.6]}} & \shortstack{\textbf{85.7}\\{\scriptsize [81.0--90.0]}} & \shortstack{\textbf{81.0}\\{\scriptsize [75.8--85.7]}} & \shortstack{\textbf{82.8}\\{\scriptsize [77.9--87.4]}} \\ \bottomrule
\end{tabular}%

\end{table}

\subsection*{WSI-Relevance Evaluation}

Table~\ref{tab:reader_ndcg} records the nDCG@5 values displayed in Figure~\ref{HiSRA_reader_study}(a).

\begin{table*}[ht!]
\centering
\caption{\textbf{The calculation uses each reader's 0--3 relevance score once per unique WSI and a query-specific ideal discounted cumulative gain at rank 5 (IDCG@5) formed from the five highest-rated slides in the full candidate pool for ranking quality evaluation.} nDCG@5 values are reported as percentages on a 0--100 scale.}
\label{tab:reader_ndcg}
\begin{tabular}{llcccc}
\toprule
\textbf{Dataset} & \textbf{Reader} & \textbf{CONCH} & \textbf{CHIEF} & \textbf{SISH} & \textbf{PathSearch} \\
\midrule
\multirow{5}{*}{HCC grading}
& Reader 1 & 78.6 & 86.7 & 82.8 & 88.0       \\
& Reader 2 & 84.8 & 91.3 & 86.0  & 93.3      \\
& Reader 3 & 93.5 & 95.0  & 93.8 & 96.8      \\
& Reader 4 & 87.5 & 92.6 & 89.0  & 95.6      \\
& Mean     & 86.1 & 91.4 & 87.9 & 93.4     \\  
\midrule
\multirow{5}{*}{Camelyon17}
& Reader 1 & 70.3  & 70.1 & 71.8 & 76.7      \\
& Reader 2 & 67.3  & 74.0  & 74.8 & 78.8      \\
& Reader 3 & 83.1  & 86.2 & 86.8 & 93.5      \\
& Reader 4 & 74.3  & 77.7 & 82.6 & 84.6      \\
& Mean     & 73.8 & 77.0  & 79.0  & 83.4    \\ 
\bottomrule
\end{tabular}%

\end{table*}

\subsection*{Attentive Mosaic Ablation Results}

Table~\ref{tab:mosaic_ablation} reports the paired I2I performance with and without the attentive mosaic across all eight tasks; the PathSearch w/o mosaic values are reconstructed from the PathSearch results and the relative changes displayed in Figure~\ref{parameter_study}(a).

\begin{table*}[ht!]
\centering
\caption{\textbf{Image-to-image retrieval performance with and without the attentive mosaic across the eight primary evaluation tasks.} PathSearch w/o mosaic excludes the attentive mosaic, whereas PathSearch includes it.}
\label{tab:mosaic_ablation}
\begin{tabular}{llrrr}
\toprule
\textbf{Dataset} & \textbf{Model variant} & \textbf{Acc@1 (\%)} & \textbf{MV@3 (\%)} & \textbf{MV@5 (\%)} \\
\midrule
\multirow{2}{*}{TCGA} & PathSearch w/o mosaic & 90.58 & 91.42 & 92.37 \\
 & PathSearch & 91.40 & 91.60 & 92.00 \\
\addlinespace
\multirow{2}{*}{H1 PCa} & PathSearch w/o mosaic & 82.74 & 84.78 & 84.51 \\
 & PathSearch & 83.90 & 85.80 & 86.20 \\
\addlinespace
\multirow{2}{*}{H1 Frozen LCa} & PathSearch w/o mosaic & 80.21 & 82.93 & 83.71 \\
 & PathSearch & 82.30 & 85.00 & 85.80 \\
\addlinespace
\multirow{2}{*}{Camelyon17} & PathSearch w/o mosaic & 81.63 & 83.98 & 83.27 \\
 & PathSearch & 81.30 & 83.90 & 84.10 \\
\addlinespace
\multirow{2}{*}{UBC-OCEAN} & PathSearch w/o mosaic & 75.80 & 79.47 & 81.64 \\
 & PathSearch & 75.80 & 80.50 & 82.70 \\
\addlinespace
\multirow{2}{*}{BRACS} & PathSearch w/o mosaic & 73.83 & 79.31 & 80.80 \\
 & PathSearch & 75.90 & 80.10 & 81.20 \\
\addlinespace
\multirow{2}{*}{H1+H2 RareCancer} & PathSearch w/o mosaic & 93.48 & 95.80 & 95.12 \\
 & PathSearch & 97.50 & 98.00 & 97.50 \\
\addlinespace
\multirow{2}{*}{H3 HCC} & PathSearch w/o mosaic & 87.91 & 83.53 & 81.00 \\
 & PathSearch & 88.70 & 85.70 & 81.00 \\
\bottomrule
\end{tabular}%

\end{table*}

\begin{table}[t]
\centering
\caption{Indexing Time Analysis. Results averaged over 100 randomly selected TCGA WSIs with a batch size of 196 for tiled patch embedding. All experiments were conducted on the same device described in Section~\ref{sec_implementation_details} but using a single RTX 3090 GPU instead to better reflect realistic clinical deployment scenarios, although using a larger batch size with modern GPUs can further reduce the patch embedding time gap between embedding-first and clustering-first methods. To isolate differences arising from mosaic construction and aggregation strategies, patch embedding time for all methods was measured using CONCH. For clustering-based methods, aggregation time additionally includes the I/O overhead associated with mosaic image writing and reading. RetCCL represents an exception within the embedding-first category, as it still relies on clustering-based mosaic construction in a high-dimensional embedding space, leading to higher aggregation time and overall indexing cost.}
\label{indexing_time_comparison}
\begin{tabular}{@{}lccccc@{}}
\toprule
\multicolumn{1}{l|}{\multirow{2}{*}{\textbf{Method}}} & \multicolumn{2}{c|}{\textbf{Configuration}}                          & \multicolumn{3}{c}{\textbf{Timing Breakdown (s)}}                                                    \\ \cmidrule(l){2-6} 
\multicolumn{1}{l|}{}                                 & \textbf{Patches}       & \multicolumn{1}{c|}{\textbf{Coverage}}      & \multicolumn{1}{c|}{\textbf{Embedding}} & \multicolumn{1}{c|}{\textbf{Aggregation}} & \textbf{Total Time} \\ \midrule
\multicolumn{6}{l}{\textit{\textbf{Embedding-First Paradigm}}}                                                                \\ \midrule
\multicolumn{1}{l|}{Mean Pooling Aggregator}                  & \multirow{5}{*}{2,755} & \multicolumn{1}{c|}{\multirow{5}{*}{100\%}} & \multicolumn{1}{c|}{\multirow{5}{*}{120.41}}  & \multicolumn{1}{c|}{1e-4}               & 120.41         \\
\multicolumn{1}{l|}{RetCCL Aggregator}                           &                        & \multicolumn{1}{c|}{}                       & \multicolumn{1}{c|}{}                         & \multicolumn{1}{c|}{12.44}                     &  132.85              \\
\multicolumn{1}{l|}{GigaPath Aggregator}                            &                        & \multicolumn{1}{c|}{}                       & \multicolumn{1}{c|}{}                         & \multicolumn{1}{c|}{0.21}               & 120.62         \\
\multicolumn{1}{l|}{CHIEF Aggregator}                            &                        & \multicolumn{1}{c|}{}                       & \multicolumn{1}{c|}{}                         & \multicolumn{1}{c|}{3e-3}               & 120.41         \\
\multicolumn{1}{l|}{\textbf{PathSearch}}                       &                        & \multicolumn{1}{c|}{}                       & \multicolumn{1}{c|}{}                         & \multicolumn{1}{c|}{\textbf{1e-2}}               & \textbf{120.42}         \\ \midrule
\multicolumn{6}{l}{\textit{\textbf{Clustering-First Paradigm}}}                                                                                                                                                                           \\ \midrule
\multicolumn{1}{l|}{Yottixel / SISH}                  & 275                    & \multicolumn{1}{c|}{10\%}                   & \multicolumn{1}{c|}{37.76}                    & \multicolumn{1}{c|}{9.75}                 & 47.51          \\ \bottomrule
\end{tabular}
\end{table}

\begin{table}[htbp]
\centering
\caption{\textbf{Training sample distribution across TCGA projects.} A breakdown of the 6,926 slide-report pairs used for training the PathSearch model, categorized by their respective TCGA project study name.}
\label{tcga_training}
\begin{tabular}{@{}llr@{}}
\toprule
\textbf{Project ID} & \textbf{Cancer Type} & \textbf{Samples} \\
\midrule
TCGA-BRCA & Breast invasive carcinoma & 684 \\
TCGA-THCA & Thyroid carcinoma & 479 \\
TCGA-UCEC & Uterine corpus endometrial carcinoma & 463 \\
TCGA-HNSC & Head and neck squamous cell carcinoma & 424 \\
TCGA-COAD & Colon adenocarcinoma & 399 \\
TCGA-PRAD & Prostate adenocarcinoma & 358 \\
TCGA-KIRC & Kidney renal clear cell carcinoma & 349 \\
TCGA-LGG  & Brain lower-grade glioma & 339 \\
TCGA-BLCA & Bladder urothelial carcinoma & 327 \\
TCGA-LIHC & Liver hepatocellular carcinoma & 315 \\
TCGA-LUSC & Lung squamous cell carcinoma & 310 \\
TCGA-LUAD & Lung adenocarcinoma & 293 \\
TCGA-STAD & Stomach adenocarcinoma & 288 \\
TCGA-CESC & Cervical squamous cell carcinoma and endocervical adenocarcinoma & 246 \\
TCGA-KIRP & Kidney renal papillary cell carcinoma & 173 \\
TCGA-PAAD & Pancreatic adenocarcinoma & 162 \\
TCGA-PCPG & Pheochromocytoma and paraganglioma & 155 \\
TCGA-READ & Rectum adenocarcinoma & 152 \\
TCGA-GBM  & Glioblastoma multiforme & 134 \\
TCGA-SARC & Sarcoma & 126 \\
TCGA-ESCA & Esophageal carcinoma & 114 \\
TCGA-THYM & Thymoma & 107 \\
TCGA-SKCM & Skin cutaneous melanoma & 94 \\
TCGA-TGCT & Testicular germ cell tumors & 72 \\
TCGA-KICH & Kidney chromophobe & 69 \\
TCGA-UVM  & Uveal melanoma & 65 \\
TCGA-OV   & Ovarian serous cystadenocarcinoma & 61 \\
TCGA-MESO & Mesothelioma & 60 \\
TCGA-UCS  & Uterine carcinosarcoma & 39 \\
TCGA-CHOL & Cholangiocarcinoma & 34 \\
TCGA-DLBC & Lymphoid neoplasm diffuse large B-cell lymphoma & 25 \\
TCGA-ACC  & Adrenocortical carcinoma & 10 \\
\midrule
\textbf{Total} & & \textbf{6926} \\
\bottomrule
\end{tabular}
\end{table}

\begin{table}[t]
\centering
\caption{\textbf{Detailed sample distribution across TCGA projects for internal testing.} The internal test set comprises 502 slides/patients sampled from six TCGA projects for organ-level and subtype-level retrieval evaluation.}
\label{tcga_testing}
\begin{tabular}{l l r}
\toprule
\textbf{TCGA Project} & \textbf{Study Name} & \textbf{Samples} \\
\midrule
TCGA-BRCA & Breast invasive carcinoma & 167 \\
TCGA-KIRC & Kidney renal clear cell carcinoma & 99 \\
TCGA-LUSC & Lung squamous cell carcinoma & 82 \\
TCGA-LUAD & Lung adenocarcinoma & 86 \\
TCGA-KIRP & Kidney renal papillary cell carcinoma & 50 \\
TCGA-KICH & Kidney chromophobe renal cell carcinoma & 18 \\
\midrule
\textbf{Total} & & \textbf{502} \\
\bottomrule
\end{tabular}
\end{table}






\end{document}